\def\year{2022}\relax
\documentclass[letterpaper]{article} 
\usepackage{aaai22}  
\usepackage{times}  
\usepackage{helvet}  
\usepackage{courier}  
\usepackage[hyphens]{url}  
\usepackage{graphicx} 
\usepackage{natbib}  
\usepackage{caption} 
\DeclareCaptionStyle{ruled}{labelfont=normalfont,labelsep=colon,strut=off} 
\usepackage{algorithm}
\usepackage{algorithmic}
\usepackage{amsmath}
\usepackage{graphicx}
\usepackage{caption}
\usepackage{subcaption}
\usepackage{amssymb}

\usepackage{textcomp}
\usepackage{booktabs}

\usepackage{newfloat}
\usepackage{listings}
\floatstyle{ruled}
\newfloat{listing}{tb}{lst}{}
\floatname{listing}{Listing}
\title{Sample-Efficient Reinforcement Learning via \\ Conservative Model-Based Actor-Critic}
\author{
Zhihai Wang\textsuperscript{\rm 1}, 
Jie Wang\textsuperscript{\rm 1,2}\thanks{Corresponding author.}, 
Qi Zhou\textsuperscript{\rm 1},
Bin Li\textsuperscript{\rm 1}, 
Houqiang Li\textsuperscript{\rm 1,2}
}

\affiliations{
\textsuperscript{\rm 1}CAS Key Laboratory of Technology in GIPAS, 
University of Science and Technology of China\\
\textsuperscript{\rm 2}Institute of Artificial Intelligence, Hefei Comprehensive National Science Center\\
\{zhwangx, zhouqida\}@mail.ustc.edu.cn\\
\{jiewangx, binli, lihq\}@ustc.edu.cn
}

\begin{document}

\maketitle

\begin{abstract} 

    Model-based reinforcement learning algorithms, which aim to learn a model of the environment to make decisions, are more sample efficient than their model-free counterparts. The sample efficiency of model-based approaches relies on whether the model can well approximate the environment. However, learning an accurate model is challenging, especially in complex and noisy environments. To tackle this problem, we propose the \textbf{c}onservative \textbf{m}odel-\textbf{b}ased \textbf{a}ctor-\textbf{c}ritic (CMBAC), a novel approach that achieves high sample efficiency without the strong reliance on accurate learned models. Specifically, CMBAC learns multiple estimates of the Q-value function from a set of inaccurate models and uses the average of the bottom-k estimates---a conservative estimate---to optimize the policy. An appealing feature of CMBAC is that the conservative estimates effectively encourage the agent to avoid unreliable ``promising actions''---whose values are high in only a small fraction of the models. Experiments demonstrate that CMBAC significantly outperforms state-of-the-art approaches in terms of sample efficiency on several challenging tasks, and the proposed method is more robust than previous methods in noisy environments.

\end{abstract}

\section{Introduction}
    Reinforcement learning has achieved great success in decision-making tasks, ranging from playing video games \cite{nature_dqn, hessel2018rainbow} to controlling robots in simulators \cite{lillicrap2015continuous,sac}. However, many of these results are achieved by model-free algorithms and generally require a massive number of samples, which significantly hinders the applications of model-free methods in real-world tasks \cite{kurutach2018metrpo, janner2019mbpo}. In contrast, model-based approaches, which build a model of the environment and generate fictitious interactions, are more sample-efficient than their model-free counterparts. Therefore, model-based approaches are promising candidates for dealing with real-world tasks. 

    The sample efficiency of model-based approaches crucially relies on learning accurate models efficiently, as model errors limit the performance of learned policies, known as the model-bias problem \cite{deisenroth2011pilco, mb_mpo}. Specifically, model errors can mislead the agent into selecting the unreliable actions---whose values are high in the model but is low in the true environment with a large probability---and thus degrade the performance of the learned policy until learning an accurate model from a large number of interactions \cite{kurutach2018metrpo, mb_mpo}. Previous methods tackle this problem by improving the expressive power of the models, such as using neural networks and ensemble models \cite{PunjaniA15, NagabandiKFL18, kurutach2018metrpo, chua2018pets}. However, efficiently learning accurate models remains challenging, especially in complex and noisy environments \cite{janner2019mbpo, benchmark, m2ac}, which creates a hindrance in further improving the sample efficiency of model-based approaches.

    To alleviate the strong reliance on accurate models, recent model-based algorithms \cite{slbo, pombu, mopo, m2ac} prevent the agent from exploiting model errors by an uncertainty-based penalty. Specifically, they explicitly quantify the uncertainty of the Q-value via the discrepancy of ensemble models and use it as a penalty to learn a lower bound of the true Q-value. However, many works have shown that the uncertainty quantification of existing methods can be unreliable, which acts as a bottleneck to their sample efficiency \cite{trust_uncertainty, combo}. We also empirically show that many existing uncertainty quantification methods can not well approximate the errors of Q-function in Section \ref{sec:visualization}.  
    
    In this paper, we propose the
    \textbf{c}onservative \textbf{m}odel-\textbf{b}ased  \textbf{a}ctor-\textbf{c}ritic
    (CMBAC), a novel approach that approximates a posterior distribution over Q-values based on the ensemble models and uses the average of the left tail of the distribution approximation to optimize the policy. 
    Specifically, CMBAC alternates between learning multiple estimates of the Q-value from the ensemble models and uses a conservative estimate, i.e., the average of the bottom-k estimates, to optimize the policy. CMBAC has two main advantages: (1) the conservative estimates effectively encourage the agent to avoid unreliable ``promising actions''---whose values are high in only a small fraction of the models; (2) the distribution approximation over Q-values can produce reasonable uncertainty estimates of the Q-value.  
    We empirically show that the uncertainty quantification of CMBAC approximates the errors of Q-function more accurately than previous uncertainty quantification methods, which plays a crucial role in the impressive performance of CMBAC (please refer to Figure \ref{fig:ablation_component}). Experiments show that CMBAC significantly outperforms state-of-the-art methods in terms of sample efficiency on several challenging control tasks \cite{openaigym, mujoco}. Moreover, experiments demonstrate that CMBAC is more robust to model imperfections than previous methods in noisy environments. 

\section{Related Work}
    In this section, we discuss related work, including model-based reinforcement learning, uncertainty in reinforcement learning, and conservatism in reinforcement learning. 
    
    \subsection{Model-Based Reinforcement Learning}
        Roughly speaking, model-based approaches fall into three categories according to the way of model usage: (1) dyna-style methods \cite{sutton_dyna_style, slbo, pombu, janner2019mbpo}, which use the model to generate imaginary samples as additional training data; (2) shooting algorithms \cite{cem, chua2018pets, planning_with_net}, which use the model to plan to seek the optimal action sequence;
        (3) policy search with backpropagation \cite{nguyen1990neural, vgl, svg, maac, sac_svg} through time, which exploits the model derivatives and computes the analytic policy gradient. Our work falls into the first category, i.e., the dyna-style algorithm, which has recently shown the potential to achieve high sample efficiency \cite{janner2019mbpo}. 
        
    \subsection{Uncertainty in Reinforcement Learning}
        Uncertainty estimation plays a crucial role in many reinforcement learning methods \cite{mbieeb,bootstrap_dqn,ube,pombu,mopo}.  
        In model-based reinforcement learning, recent methods \cite{slbo, mopo, m2ac} mainly focus on explicitly quantifying the uncertainty of the Q-value via the discrepancy of the ensemble models to alleviate the strong reliance on accurate models. However, their uncertainty quantification can be unreliable for deep network models \cite{trust_uncertainty, combo}. In contrast, our work approximates a posterior distribution over Q-values based on the ensemble models to capture the uncertainty of the value function. Although the work \cite{uambpo} also approximates a posterior distribution over Q-values based on the ensemble models, our work differs from it in the way of value estimation. \citet{uambpo} estimates the value in a model via Monte Carlo methods \cite{sutton2018reinforcement}, which can suffer from compounding model errors. 
        In contrast, we use temporal difference methods to estimate the value \cite{sutton2018reinforcement}. 
        In model-free reinforcement learning, many methods \cite{mbieeb, bootstrap_dqn, ube} quantify the uncertainty via the true environment samples and use it as a reward bonus to promote exploration, while these uncertainty quantification methods are inappropriate for model-based methods \cite{pombu}. 
        
    \subsection{Conservatism in Reinforcement Learning}
        Previous methods introduce the conservatism into policy optimization by underestimating the true value to improve the robustness \cite{robust_optimal_control, NIPS2013_robust_mdp, epopt} and alleviate the overestimation bias \cite{td3, tqc, cql}. In model-based reinforcement learning, some methods \cite{robust_optimal_control, NIPS2013_robust_mdp, epopt} leverage robust policy optimization, which learns a policy that performs well across models. However, the learned policies tend to be over-conservative \cite{mb_mpo}. In contrast, we use the average of the bottom-k estimates instead of the minimum to optimize the policy, controlling the degree of conservatism. In model-free reinforcement learning, many methods incorporate conservatism into policy learning to alleviate the overestimation bias that comes from function approximation errors \cite{td3, tqc, cql}. In contrast, CMBAC leverages conservatism to alleviate the overestimation that comes from model errors (please refer to Section \ref{sec:visualization}). 
        
\section{Background}

    In this section, we present the notation and provide a brief introduction to the state-of-the-art model-based algorithm, i.e., Model-Based Policy Optimization \cite{janner2019mbpo}. 

\subsection{Preliminaries}

    We here introduce notation which we will use throughout the paper. We consider an infinite-horizon Markov decision process (MDP) denoted by a tuple $(\mathcal{S}, \mathcal{A}, P^*, r, \gamma, \rho_0)$, 
    where $\mathcal{S}$ and $\mathcal{A}$ are the sets of states and actions, respectively, $P^*: \mathcal{S} \times \mathcal{A} \times \mathcal{S} \to [0,\infty)$ is the transition probability density function with $P^*(\cdot|s,a)$ representing the conditional distribution of the next state given the current state $s$ and action $a$, $r: \mathcal{S}\times \mathcal{A} \to \mathbb{R}$ is the reward function, $\rho_0:\mathcal{S} \to [0,\infty)$ is the starting state distribution, and $\gamma$ is the discount factor. 
    Let $\pi:\mathcal{S} \to \mathcal{P}(\mathcal{A})$ be a stationary policy, where $\mathcal{P}(\mathcal{A})$ is a set of probability distribution over $\mathcal{A}$. Let $\pi(\cdot|s)$ denote the probability distribution over $\mathcal{A}$ at state $s$. In model-based reinforcement learning, we  learn a dynamics model $\hat{P}(\cdot|s,a)$ using data collected from interaction with the true MDP. For simplicity, we assume that the reward function $r(s,a)$ is known throughout the paper, but in practice, we learn a reward function. Let $S_0$ be the random variable for the initial state. Let $Q^{\pi,P}$ be the state-action value function on the model $P$ and policy $\pi$ defined by:
    \begin{align*}
        Q^{\pi,P}(s,a) = \mathbb{E}_{\pi,P}[\sum_{t=0}^{\infty} \gamma^t r(S_t,A_t)| S_0=s,A_0=a].
    \end{align*} 
    We define $\eta(\pi,P) = \mathbb{E}_{\pi}[Q^{\pi,P}(S_0,A_0)]$ as the expected reward-to-go. Our goal is to maximize the reward-to-go on the true model, that is, $\eta(\pi,P^*)$, over the policy $\pi$. 
    
\subsection{Model-Based Policy Optimization}

Model-based policy optimization (MBPO) is a state-of-the-art model-based algorithm that has achieved impressive performance \cite{janner2019mbpo}. MBPO has three ingredients: 
\begin{enumerate}
    \item \textbf{Ensemble models} MBPO trains a bootstrap ensemble of dynamics models via maximum likelihood technique \cite{chua2018pets} on dataset $\mathcal{D}_{\text{env}}$ collected from interactions with the true environment. Each member of the set is modeled as a Gaussian with mean and diagonal covariances given by neural networks.  Our work also learns an ensemble of probabilistic models.
    
    \item \textbf{Model usage} MBPO selects a model uniformly at random from the ensemble and generates a prediction from the selected model. To reduce compounding model errors introduced by long rollouts, MBPO generates many short rollouts as additional training dataset $\mathcal{D}_{\text{model}}$. 
    
    \item \textbf{Policy optimization} MBPO uses soft actor-critic (SAC) \cite{sac}---a state-of-the-art model-free algorithm based on the maximum entropy reinforcement learning framework---to optimize the policy. 
\end{enumerate}

\begin{algorithm}[t]
    \caption{Pseudo code for CMBAC}.
    \label{alg:CMBAC}
    \begin{algorithmic}[1]
        \STATE \textbf{Initialize} an ensemble of models $\{P_{\psi_{i}}\}_{i=1}^{N}$, environment dataset $\mathcal{D}_{\text{env}}$, and model dataset $\mathcal{D}_{\text{model}}$
        \STATE \textbf{Initialize} 
        policy $\pi_{\phi}$,  multi-head Q-network $\{Q_{\theta_j}\}_{j=1}^{K}$
        \FOR{$N$ epochs}
        \STATE Train models $\{P_{\psi_i}\}_{i=1}^{N}$ on $\mathcal{D}_{\text{env}}$
            \FOR{$E$ steps}
            \STATE Take action in environment using $\pi_{\phi}$; add to $\mathcal{D}_{\text{env}}$
                \FOR{$M$ model rollouts}
                \STATE Sample $s_t$ unifromly from $\mathcal{D}_{\text{env}}$ 
                \STATE Perform $k$-step model rollouts starting from $s_t$ using policy $\pi_{\phi}$; add to $\mathcal{D}_{\text{model}}$
                \ENDFOR
                \FOR{$G$ gradient updates}
                \STATE  Train value functions on model data: $\theta_j \leftarrow \theta_j - \lambda_{Q} \hat{\nabla}J_{Q}^j(\theta_j)$ for $j\in \{1,\dots,K\}$
                \STATE  Conservative policy optimization on model data: $\phi \leftarrow \phi - \lambda_{\pi} \hat{\nabla}J_{\pi}(\phi)$ 
                \ENDFOR
            \ENDFOR
        \ENDFOR
    \end{algorithmic}
\end{algorithm}

\section{Conservative Model-Based Actor-Critic}

\begin{figure*}[t]
    \centering
    \includegraphics[width=0.9\textwidth]{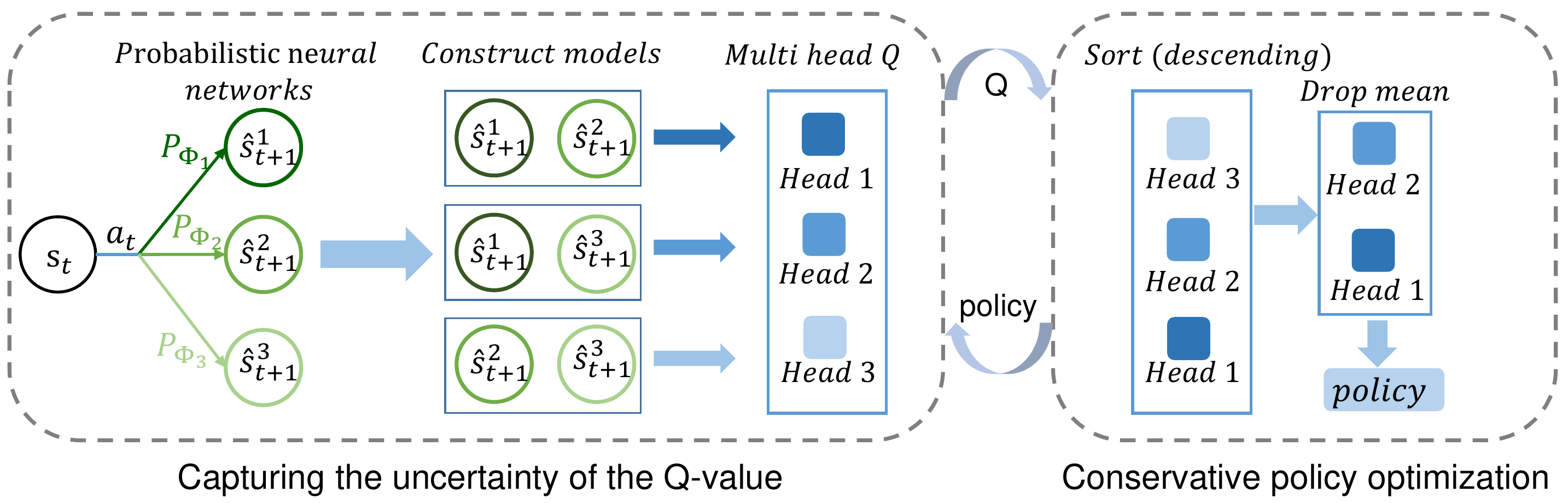}
    \caption{Illustration of CMBAC with $N=3, M=2,$ and $L=1$. It first learns three probabilistic neural networks and constructs each model using the arbitrarily two probabilistic neural networks.
    It then alternates between learning three estimates of the Q-value function from the ensemble models and using the average of the bottom-2 estimates to optimize the policy.}
    \label{fig: illustration of CMBAC}
\end{figure*}

In this section, we present a detailed description of CMBAC. CMBAC alternates between (1) learning multiple estimates of the Q-value function from the ensemble models and (2) using the average of the bottom-k estimates to optimize the policy. We provide an illustration of CMBAC in Figure \ref{fig: illustration of CMBAC} and summarize the procedure of CMBAC in Algorithm \ref{alg:CMBAC}.

\subsection{Capturing the Uncertainty of the Q-value}\label{sec:lmeq}
    To capture the uncertainty of the Q-value, CMBAC directly approximates the posterior distribution over Q-values based on the posterior distribution approximation over models. The uncertainty of the Q-value comes from the unknown true environment, and model-based approaches usually estimate it using a possible set of models. 
    
    To approximate the posterior distribution over models, CMBAC learns an ensemble of probabilistic neural networks $\{p_{\psi_i}\}_{i=1}^{N}$, which has been shown the potential to capture the uncertainty of models \cite{deep_ensemble, chua2018pets}. Each probabilistic neural network model the transition probability density as a Gaussian with mean and diagonal covariances given by neural networks. That is, $p_{\psi_i}(s^{\prime}|s,a) = \mathcal{N} (\mu_{\psi_i}(s,a), \sigma_{\psi_i}(s,a))$. To control the granularity of discrepancy between models, CMBAC constructs a set of models $\mathcal{M}$ using these probabilistic neural networks. Each element in $\mathcal{M}$, denoted by $\mathcal{M}_j$, is a set consists of $M$ different networks ($M<N$) and thus the size of $\mathcal{M}$ is $K=\tbinom{N}{M}$. We view each $\mathcal{M}_j$ as a model, and $\mathcal{M}_j$ generates next state $s^{\prime}$ given current state-action pair $(s,a)$ under the distribution
    \begin{align*}
        P_j(s^{\prime}|s,a) = \sum_{p \in (\{ p_{\psi_i}\}_{i=1}^{N}\cap \mathcal{M}_j)}\frac{1}{M} p (s^{\prime}|s,a). 
    \end{align*}
    The set of models $\mathcal{M}$ has the ability to well approximate the posterior distribution of the true environment using non-parametric bootstrap with random initialization \cite{introduction_bootstrap, bootstrap_dqn, chua2018pets}. Given fixed $M$, CMBAC achieves more fine-grained granularity of discrepancy between models in $\mathcal{M}$ by increasing the total number of networks $N$. Given fixed $N$, the discrepancy between models drops with increasing $M$, i.e., the number of networks in each $\mathcal{M}_j$. If $M=N$, then CMBAC reduces to MBPO. Moreover, representing each model by an ensemble of neural networks significantly improves the model accuracy
    \cite{kurutach2018metrpo, chua2018pets}. 

    To approximate the posterior distribution over Q-values, CMBAC learns multiple estimates via a multi-head Q-network from 
    the distribution approximation over models $\mathcal{M}$. 
    Similar to bootstrapped DQN \cite{bootstrap_dqn}, the multi-head Q-network is a shared neural network architecture with $K$ "heads" branching off independently  (please refer to Appendix D for details). Each ``head'' $\hat{Q}_{\theta_j}$ provides an estimate of the Q-value for a policy $\pi$, which corresponds to a model $\mathcal{M}_j \in \mathcal{M}$. That is, CMBAC aims to approximate the $Q^{\pi, P_j}$ via the ``head'' $\hat{Q}_{\theta_j}$. 
    
    The target value of each ``head'' $\hat{Q}_{\theta_j}(s,a)$ is given by
    \begin{align*}
        y_{j}(s,a)  =  r(s,a) + \gamma  (\hat{Q}_{\bar{\theta}_j}(s_j^{\prime}, a_j^{\prime}) - \alpha \log(\pi(a_j^{\prime}|s_j^{\prime}))),
    \end{align*}
    where $j=1,\dots, K$, $s_j^{\prime}$ is sampled from $P_j(\cdot|s,a)$, $a_j^{\prime}$ is sampled from $\pi(\cdot|s_{j}^{\prime})$, $\alpha$ is the temperature parameter, and $\hat{Q}_{\bar{\theta}_j}$ is the target value network with $\bar{\theta}_j$ being an exponentially moving average of the value network weights,
    which has been shown to stabilize training \cite{nature_dqn}.
    For each $j\in \{1,\dots, k\}$, the parameter $\theta_j$ can be trained to minimize the Bellman residual
    \begin{align*}
        J_{Q}(\theta_j)  = \mathbb{E}_{(s,a)\sim \mathcal{D}_{\text{model}}} [\frac{1}{2} (\hat{Q}_{\theta_j}(s,a) - y_{j}(s,a))^2].
    \end{align*}
    
    CMBAC naturally approximates the posterior distribution over Q-values, as it learns an estimate from a model sampled from the approximated posterior distribution over models, respectively. We use the clipped double Q-learning as proposed by \cite{td3}. In practice, we use two multi-head Q-network and train each ``head'' using the minimum of the corresponding two ``heads''.  

\subsection{Conservative Policy Optimization}
    To prevent the agent from exploiting model errors, CMBAC uses a conservative estimate of the Q-value function to optimize the policy, named conservative policy optimization. Previous model-based methods aim to learn a conservative estimate that is an approximate lower bound of the true Q-value $Q^{\pi,P^*}$. For example, robust policy optimization \cite{robust_optimal_control, NIPS2013_robust_mdp} considers a set of possible models and learns the worst-case Q-value, i.e., $Q^{\pi}(s,a) = \min_{P\in \mathcal{M}} Q^{\pi,P}(s,a)$. However, these methods tend to learn an over-conservative policy, which severely degrades their sample efficiency and asymptotic performance \cite{mb_mpo}. Unlike these methods, CMBAC introduces conservatism to alleviate the overestimation that comes from the model errors and does not aim to learn a lower bound. We observe that if we learn multiple estimates using the approach in Section \ref{sec:lmeq}, a small fraction of the “heads” severely overestimate the Q-value (see details in Section \ref{sec:visualization}) while the others provide a relatively reasonable estimation. Based on the observation, we hypothesize that the actions---whose values are high in only a small fraction of the models---are unreliable in the true environment. That is, the true Q-values of these actions are low with a large probability. Therefore, we propose to drop the top-k estimates and use the average of the others for policy optimization to alleviate the effect of such overestimation. 
    Specifically, given a state $s$ and action $a$, we sort the estimates produced by the multi-head Q-network in ascending order, denoted by $\hat{Q}_{\theta_j}(s,a)$ with $j \in [1..K]$, and then optimize the policy $\pi_{\phi}$ by minimizing the objective as proposed in \cite{sac}
    \begin{align}\label{eq:policy_optimizarion_objective}
        J_{\pi}(\phi) = \mathbb{E}_{s\sim \mathcal{D}_{\text{model}}} [\text{D}_{\text{KL}}(\pi_{\phi}(\cdot|s) \| \frac{\exp(\frac{1}{\alpha}\hat{Q} (s,\cdot))}{Z(s)})],
    \end{align}
    where
    $\hat{Q}(s,\cdot) = \frac{1}{K-L}\sum_{j=1}^{K-L} \hat{Q}_{\theta_j}(s,\cdot),$
    and $Z(s_t) = \int_{a} \exp(\frac{1}{\alpha}\hat{Q}(s,a))da$, which does not contribute to the gradient 
    with respect to the new policy.  
    
    In contrast to CMBAC, existing methods use the uncertainty of the Q-value as a penalty. However, we find it may be sensitive to the penalty coefficient (See Appendix C). 

\subsection{Discussion}
    We discuss some advantages of CMBAC in this subsection.
    
    \subsubsection{Capture global uncertainty} 
    CMBAC naturally captures the global uncertainty \cite{ube, pombu}, which considers the compounding model error and its effect on the critic learning. The global uncertainty allows CMBAC to deal with the overestimation that comes from the long-term prediction errors of the model. Otherwise, conservatively optimizing the policy via local uncertainty (i.e., the one-step prediction error) may 
    mislead the agent into exploiting the regions 
    where the one-step prediction is accurate, but multi-step prediction errors are large. 

    \subsubsection{Capture uncertainty with various granularity} CMBAC approximates a posterior distribution over Q-values to implicitly capture the uncertainty of Q-value. To achieve granularity in uncertainty capturing, CMBAC controls the granularity of distribution approximation by varying $M$, i.e., the number of neural networks in each model. The number of estimates $K$ depends on $M$, and different $K$ correspond to the different granularity of distribution approximation.
    
    \subsubsection{Flexibly control the degree of conservatism} CMBAC can flexibly control the degree of conservatism by varying $M$ and $L$. By varying $M$, CMBAC can provide fine-grained control over the degree of conservatism, as it achieves granularity in capturing uncertainty. By varying $L$, the degree of conservatism increases with $L$.
    
\begin{figure*}[t]
    \centering
    \includegraphics[width=0.25\textwidth]{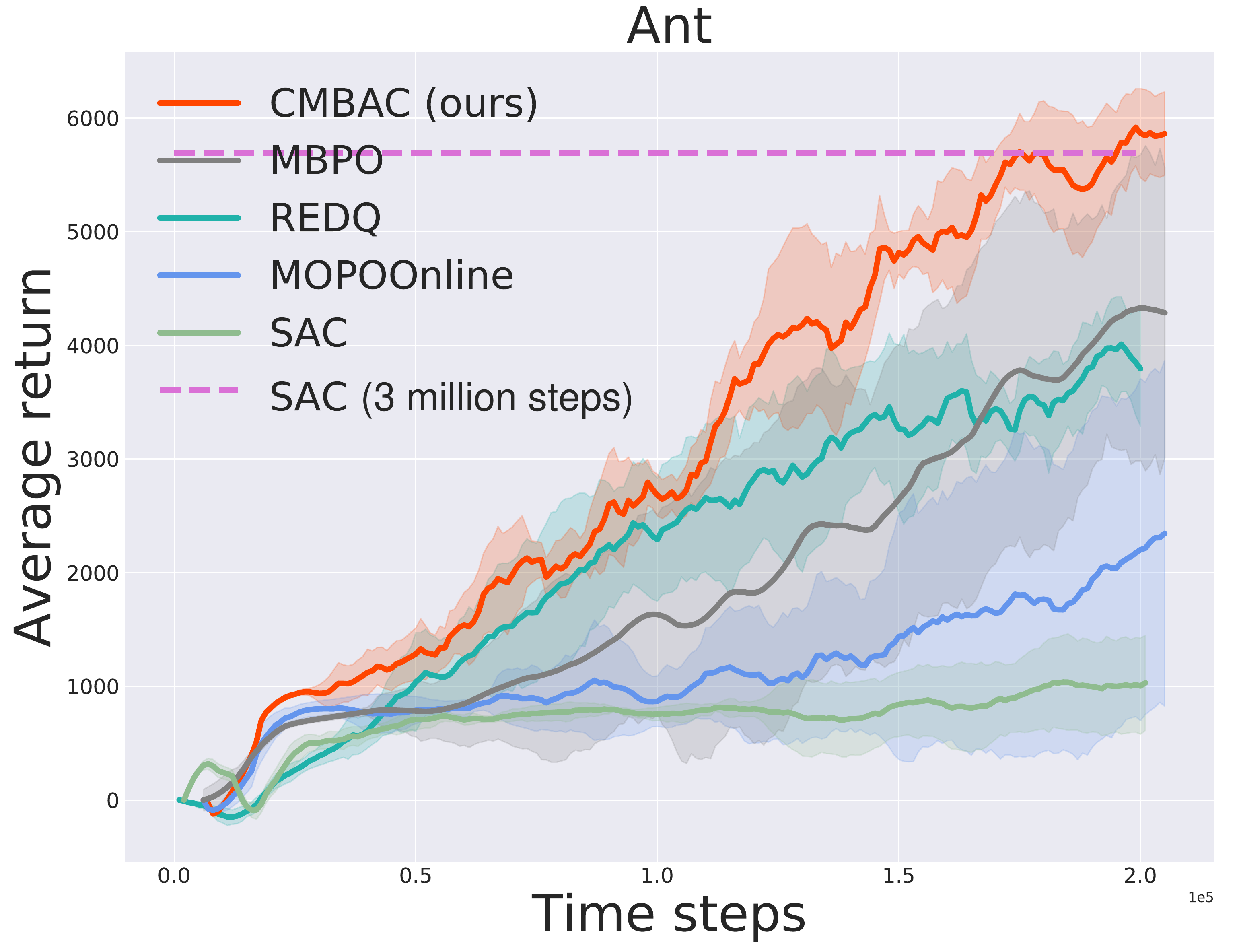}
    \includegraphics[width=0.25\textwidth]{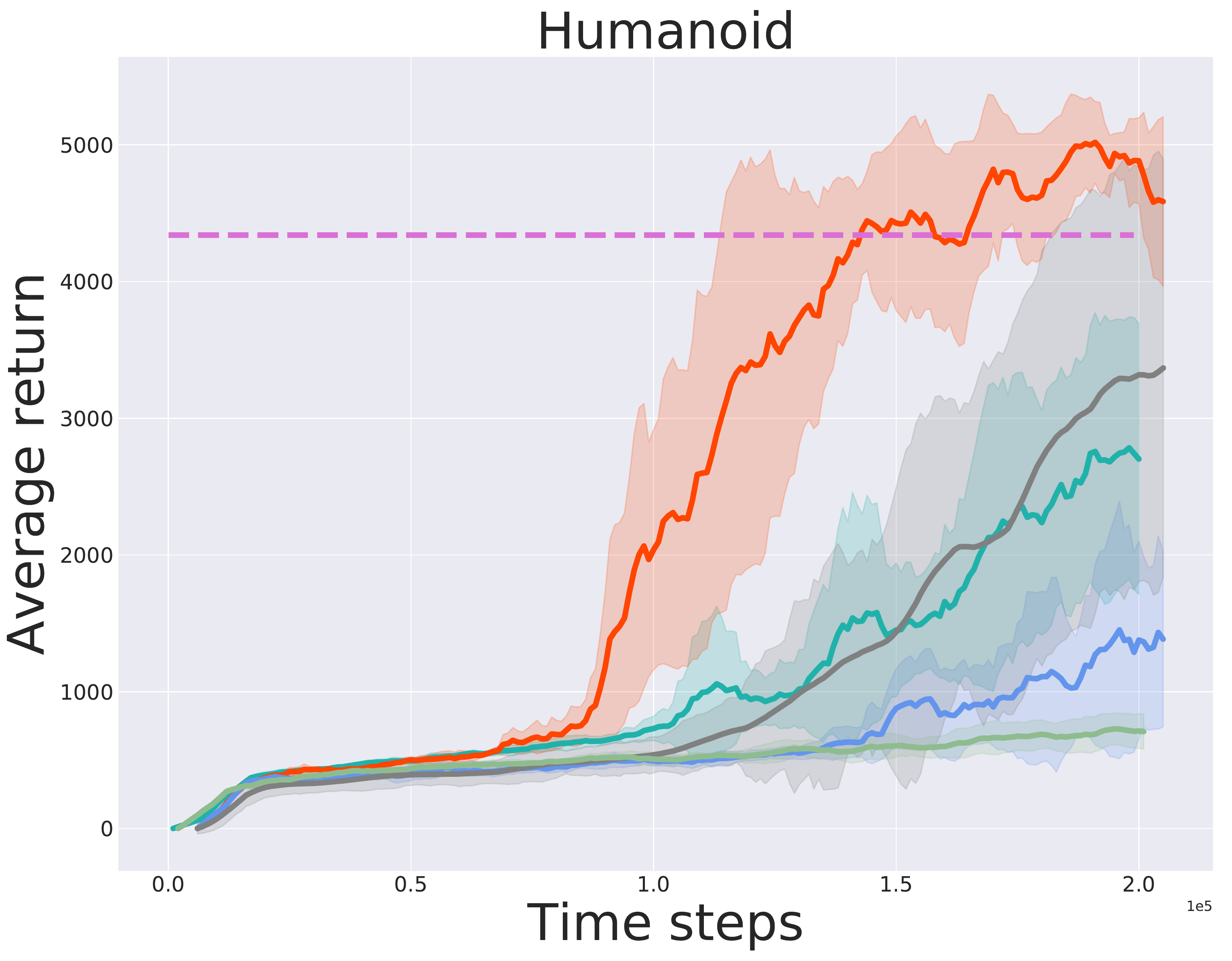}
    \includegraphics[width=0.25\textwidth]{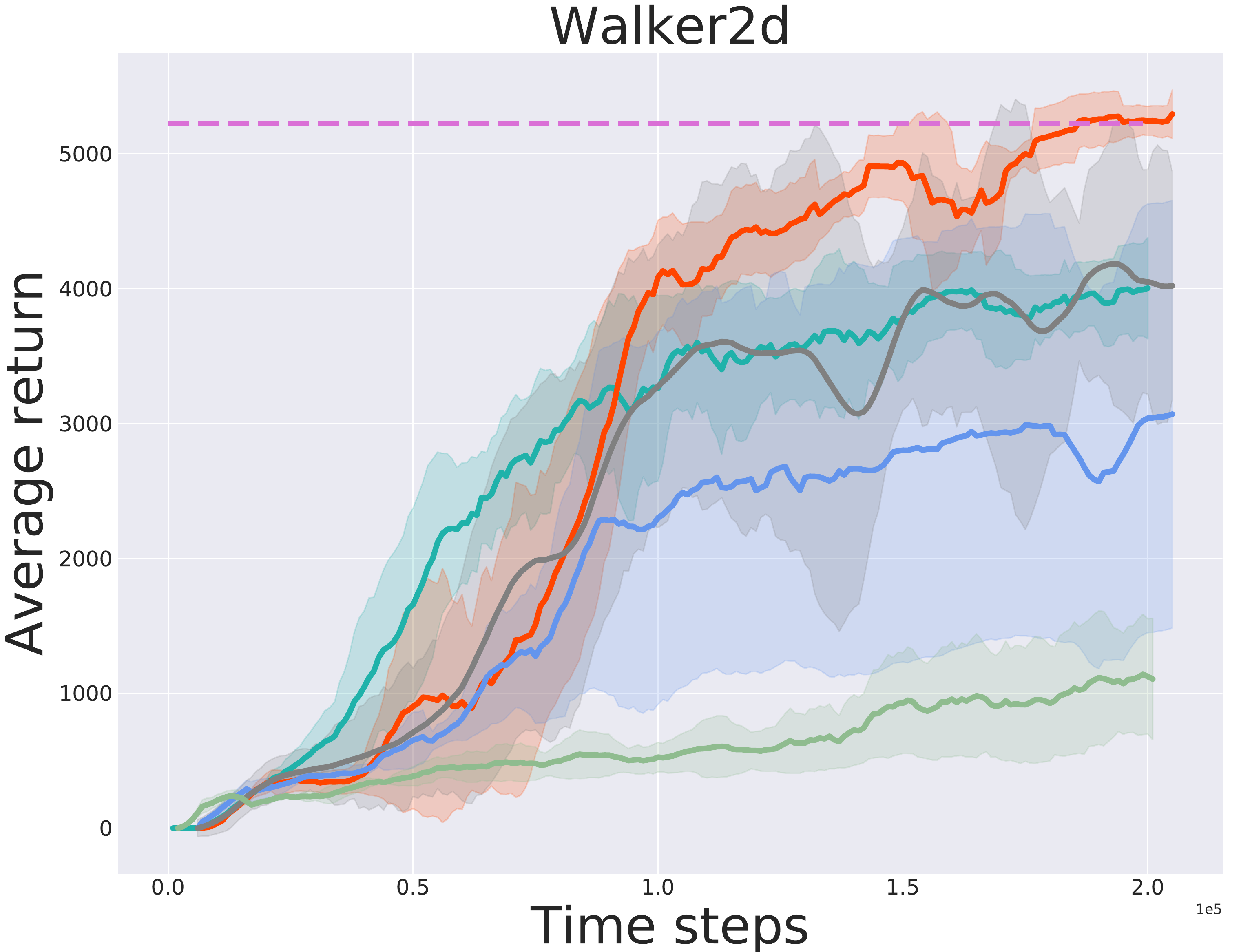}
    
    \includegraphics[width=0.25\textwidth]{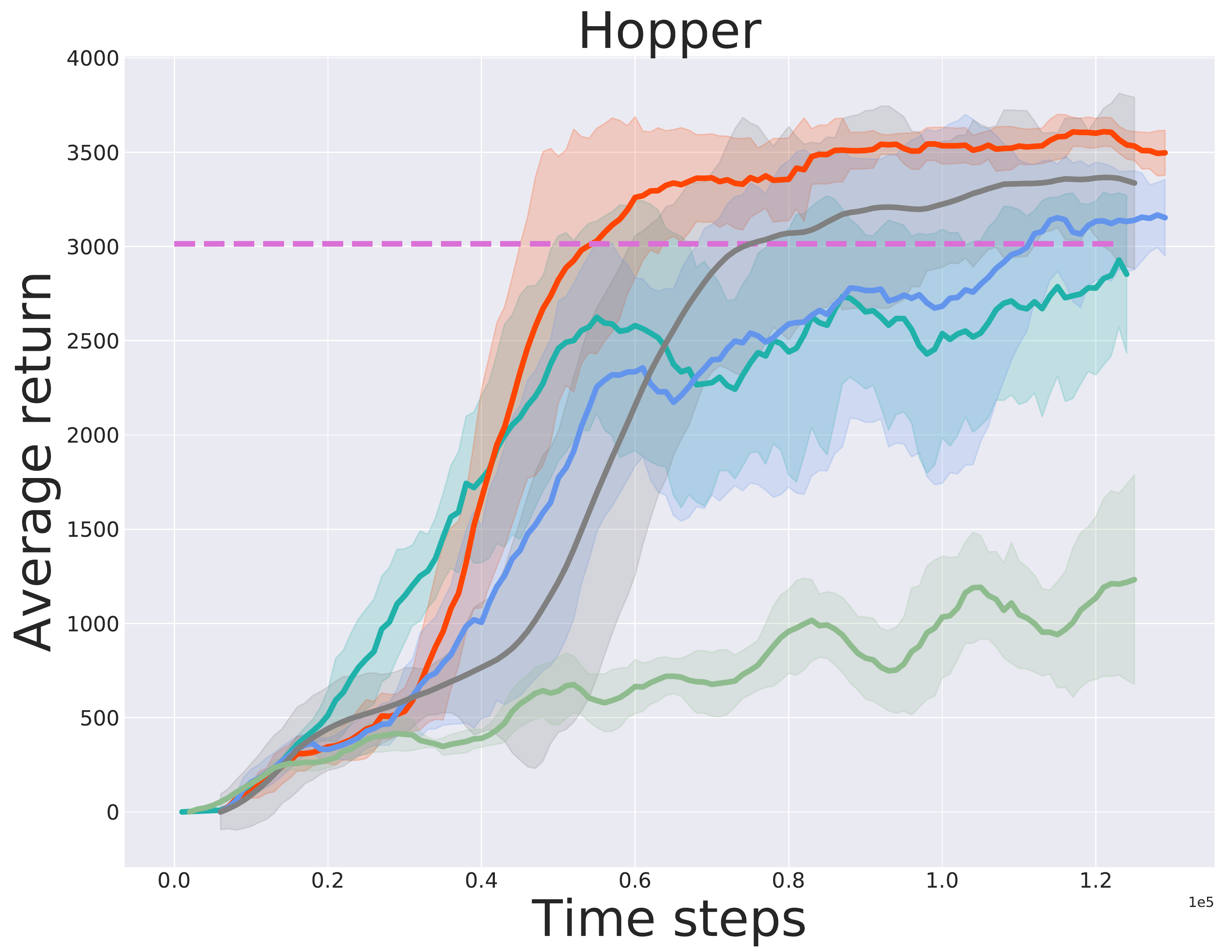}
    \includegraphics[width=0.25\textwidth]{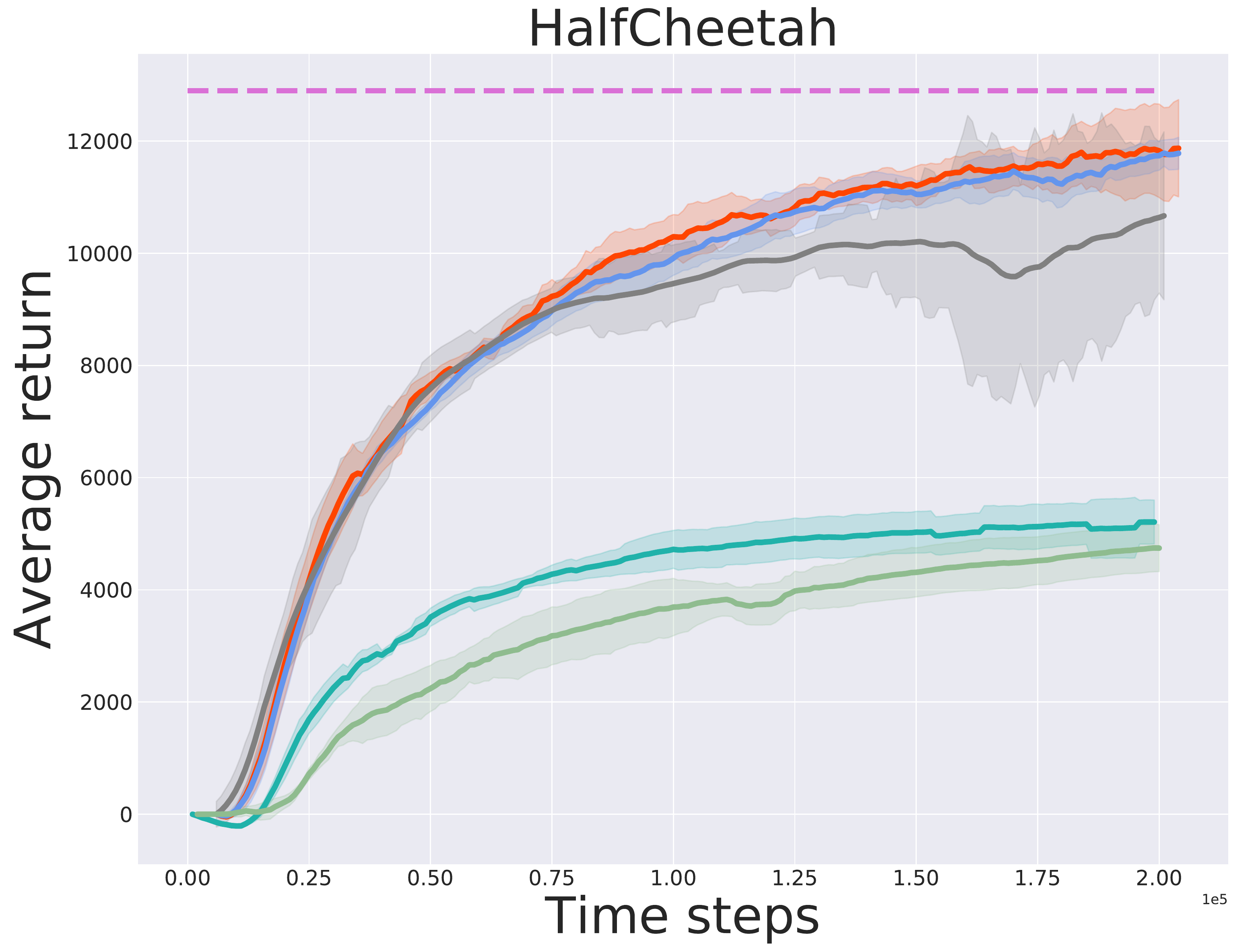}
    \includegraphics[width=0.25\textwidth]{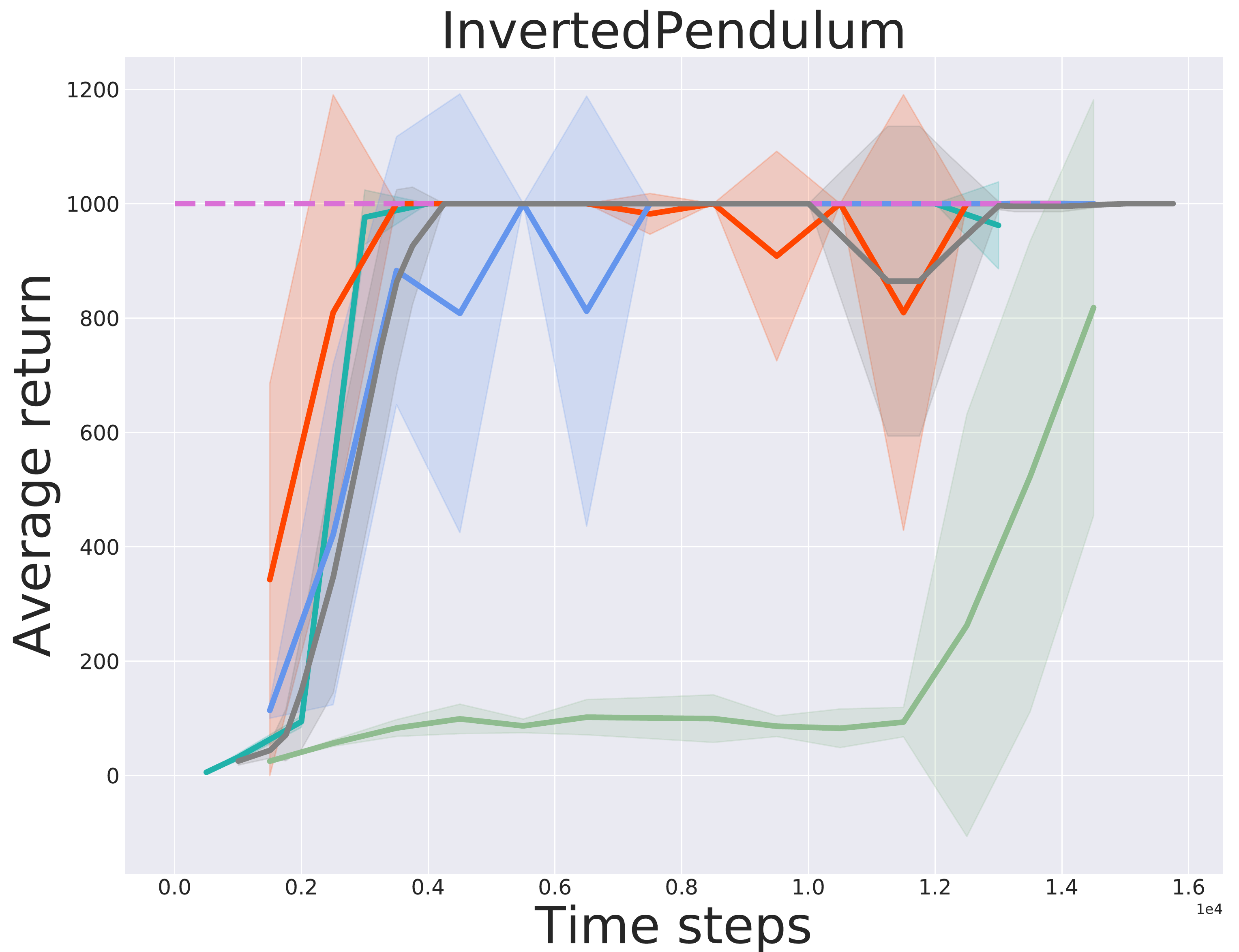}
    \caption{Performance of CMBAC and four baselines on six continuous control tasks. The solid curves correspond to the mean and the shaded region to the standard deviation over five random seeds. For visual clarity, we smooth curves uniformly. The results show that CMBAC significantly outperforms these baselines in terms of sample efficiency on several challenging tasks. }
    \label{fig: evaluation}
\end{figure*}

\begin{figure}[t] 
\centering
\includegraphics[width=0.23\textwidth]{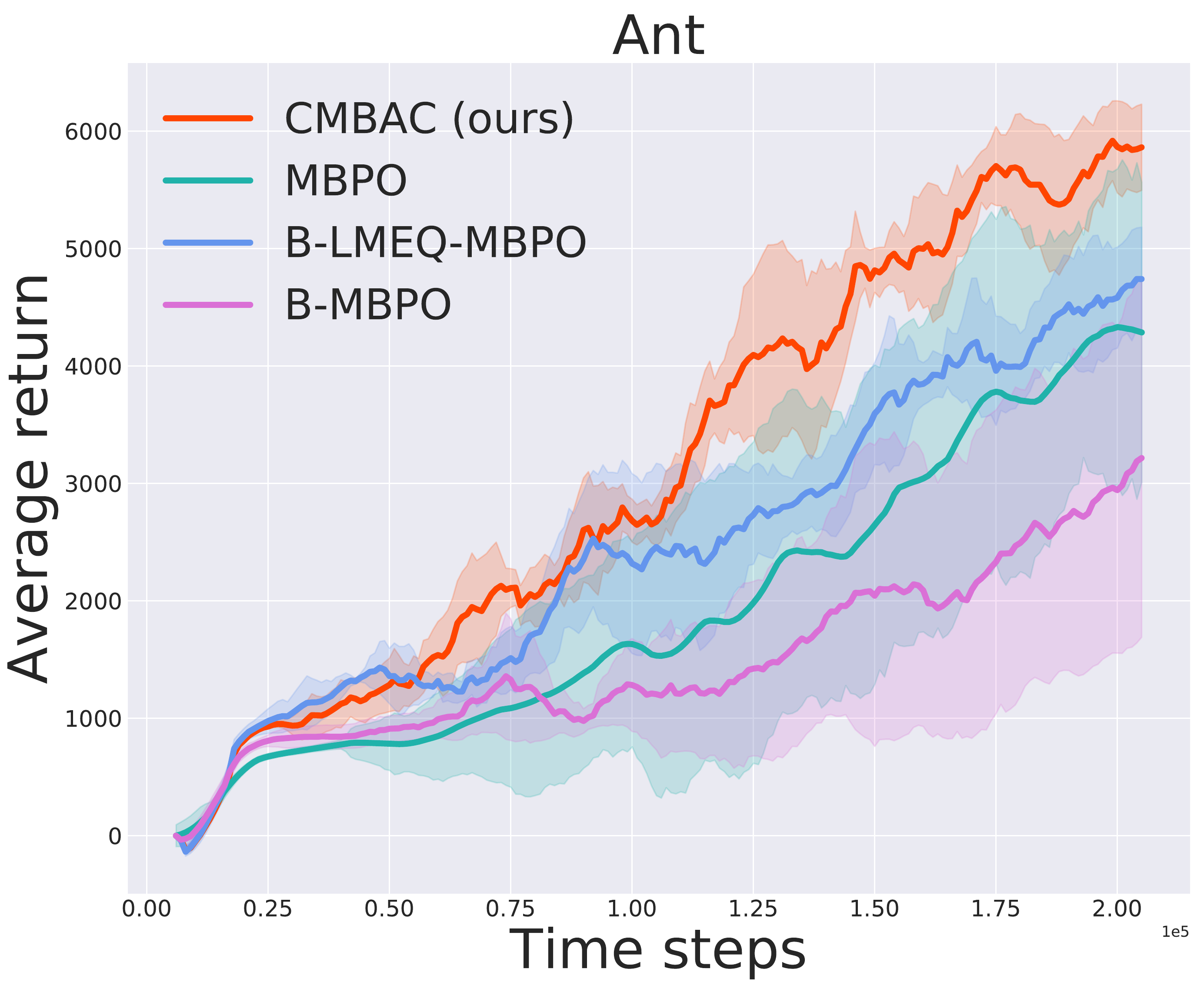}
\includegraphics[width=0.23\textwidth]{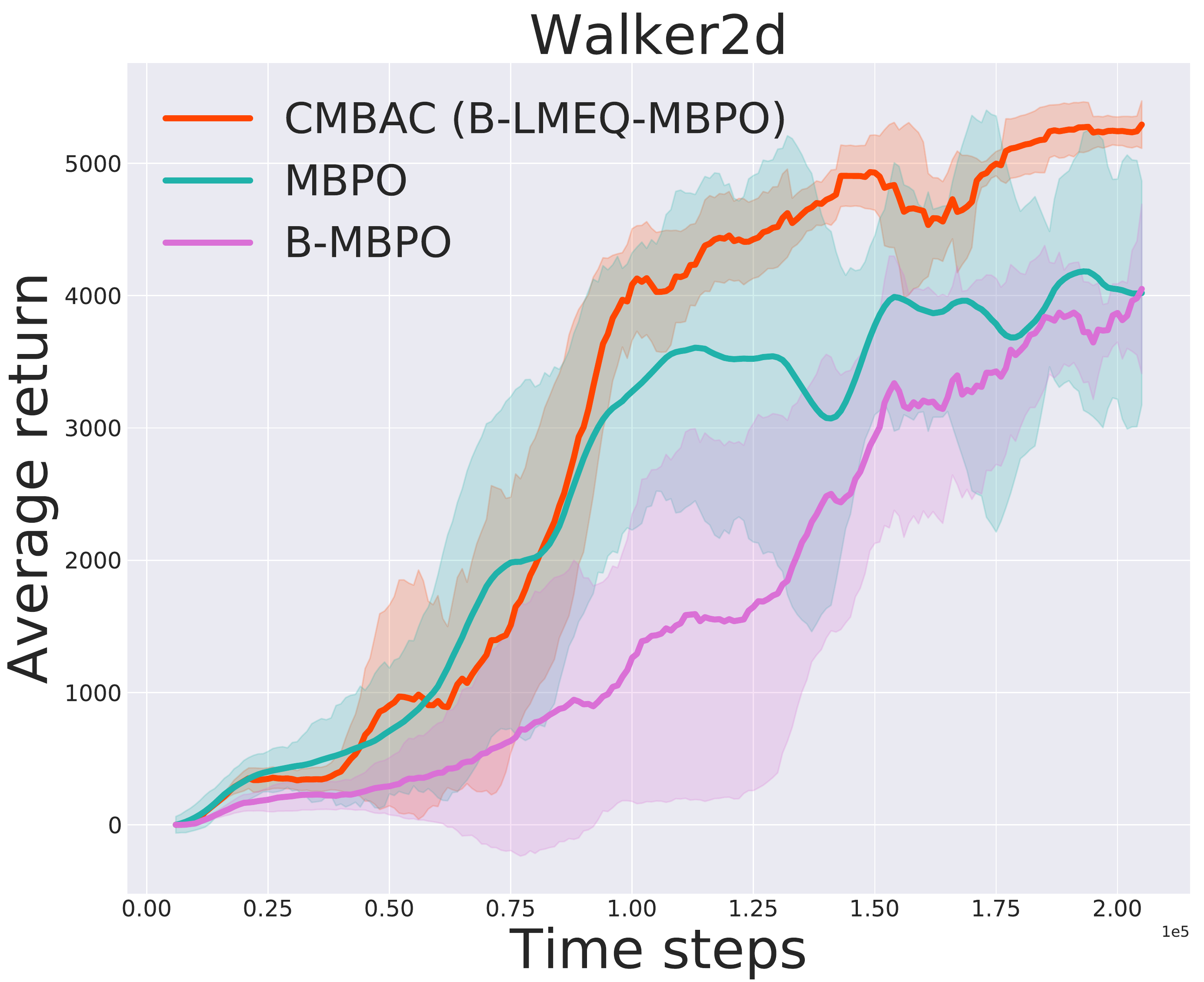}
\caption{Performance of CMBAC and its ablations on the Ant and Walker2d environments. On the Ant environment, each component of CMBAC is significant for performance improvement. On the Walker2d environment, CMBAC reduces to B-LMEQ-MBPO, as conservative policy optimization does not further improve the performance.}
\label{fig:ablation_component}
\end{figure}

\section{Experiments}
    Our experiments have four main goals in this section: (1) Test whether CMBAC can significantly outperform state-of-the-art methods. (2) Perform carefully designed ablation study of CMBAC.
    (3) Perform visualization experiments of CMBAC to explain its effectiveness. (4) Test the robustness of CMBAC in noisy environments. 

\subsection{Comparative Evaluation}
    For model-based methods, we compare our method to model-based policy optimization (MBPO) \cite{janner2019mbpo}, a state-of-the-art algorithm. In addition, we compare to an online variant of model-based offline policy optimization (MOPO) \cite{mopo}---a state-of-the-art offline model-based algorithm---which quantifies the uncertainty of the models and uses the uncertainty as a penalty for policy optimization. For a fair comparison, we implement CMBAC and MOPO-Online, i.e., the online variant of MOPO, both on top of the MBPO.
    Although masked model-based actor-critic (M2AC) \cite{m2ac} outperforms MBPO by using the uncertainty of models, its core component is a masking mechanism, which is orthogonal to our method. Therefore, we do not compare to M2AC.
    For model-free methods, we compare to soft actor-critic (SAC) \cite{sac}, which is used for policy learning in our method; randomized ensembled double Q-learning (REDQ) \cite{redq}, which has achieved comparable sample efficiency to MBPO. 
    
    We evaluate CMBAC and these baselines on MuJoCo \cite{mujoco} benchmark tasks as used in MBPO. We use two multi-head Q-networks with three hidden layers of 512 neurons each. For all environments except Walker2d, we use the number of dropped estimates $L=1$. On Walker2d, we use $L=0$. For our method, we select the hyperparameter $M$ for each environment independently via grid search. The best hyperparameter for Humanoid, Hopper, Walker2d, and the rest is $M=1, 3, 4, 2,$ respectively. The details of the experimental setup are in Appendix B.

    Figure \ref{fig: evaluation} shows that CMBAC significantly outperforms these baselines in terms of sample efficiency on several challenging control tasks. For the most challenging Humanoid environment, CMBAC learns substantially faster than state-of-the-art methods. Specifically, the performance of CMBAC on the Humanoid task at 200 thousand steps matches that of MBPO at 300 thousand steps and SAC at 3 million steps. MOPO-Online achieves poor results on several challenging tasks, which may suggest that its uncertainty-based penalty is inappropriate for the online setting (please refer to Figure \ref{fig:visulization}). 
    The model-free method REDQ has recently achieved comparable sample efficiency to MBPO, which raises the question of whether model-based methods carry the promise of being data efficient. Our results demonstrate that model-based methods can be more sample efficient than their model-free counterparts with careful model usage. 
    
    To demonstrate the hyperparameter insensitivity of CMBAC, we replot Figure \ref{fig: evaluation} using unified hyperparameters. Detailed results are in Appendix E. To demonstrate the scalability of CMBAC, we compare CMBAC and the baselines on two additional environments, i.e., Walker2d-NT and Hopper-NT \cite{benchmark} (See Appendix E).
    
\begin{figure*}[t] 
\centering
\begin{minipage}{0.48\linewidth}
    \centering
    \includegraphics[width=0.48\linewidth]{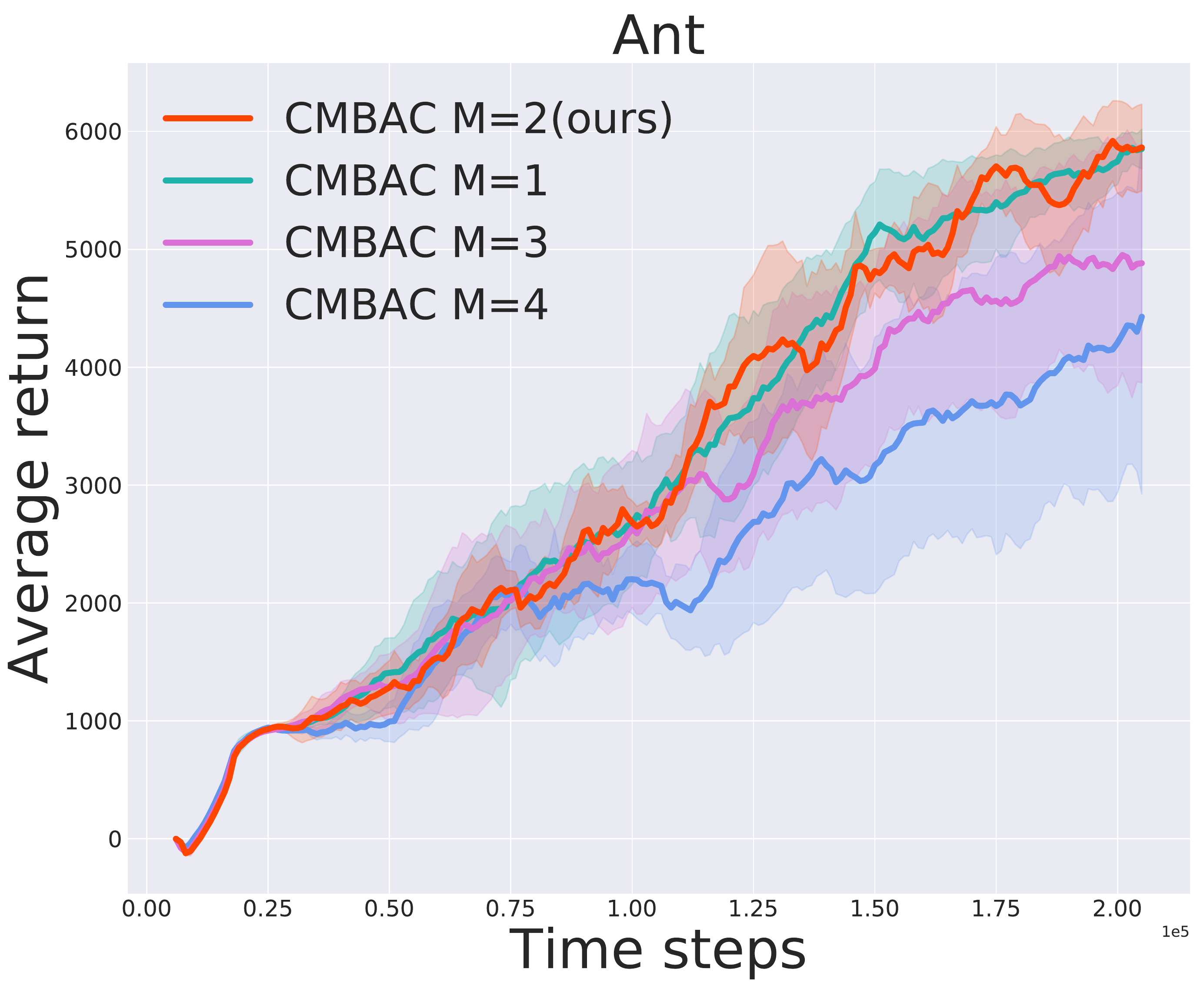}
    \includegraphics[width=0.48\linewidth]{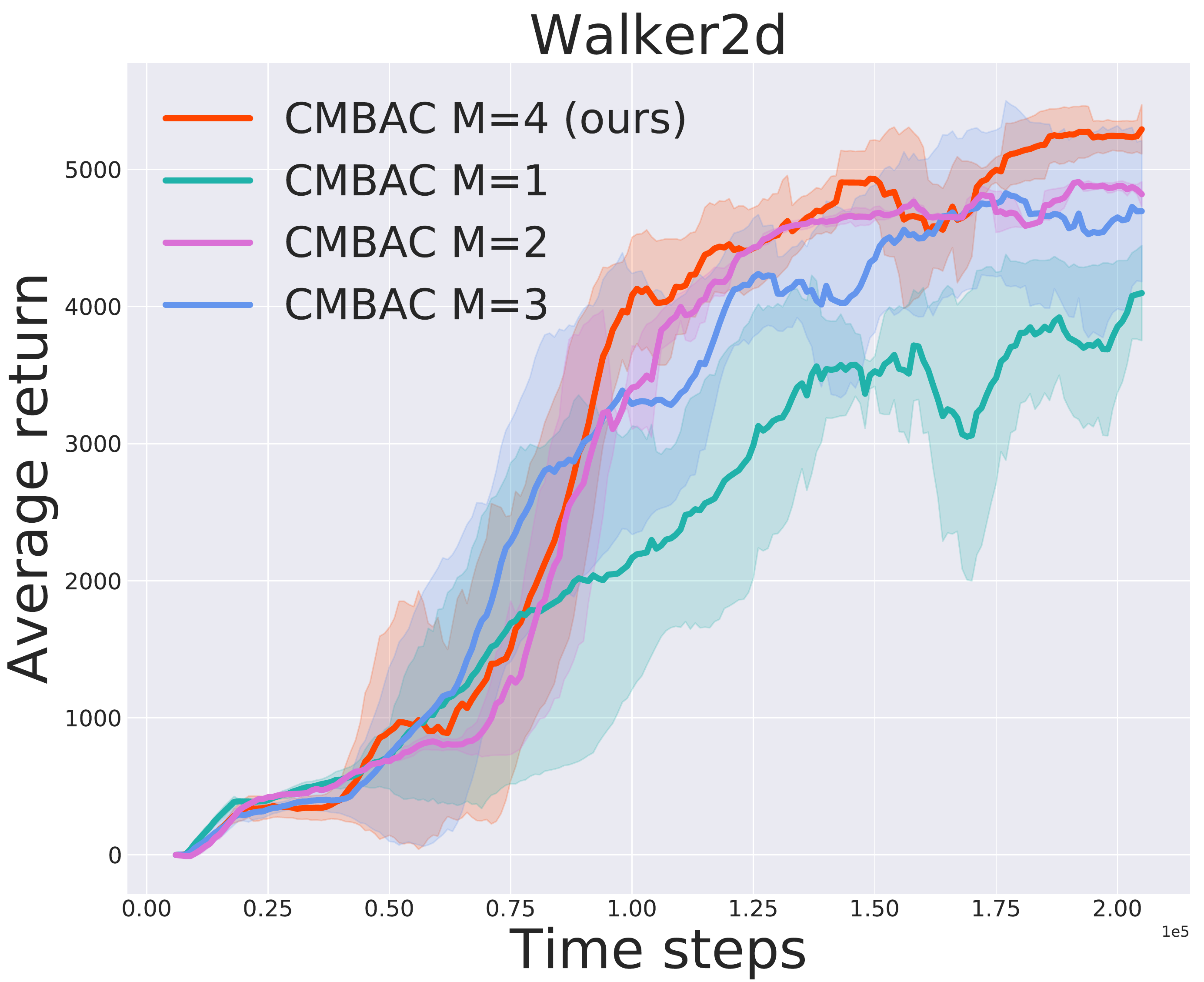}
    \caption{Varying the number of neural networks in a model $M$. The results demonstrate that $M$ is essential, and there is an optimal number, i.e., $M=2$.}
    \label{fig:ablation_subset_size}
\end{minipage}
\begin{minipage}{0.48\linewidth}
    \includegraphics[width=0.48\linewidth]{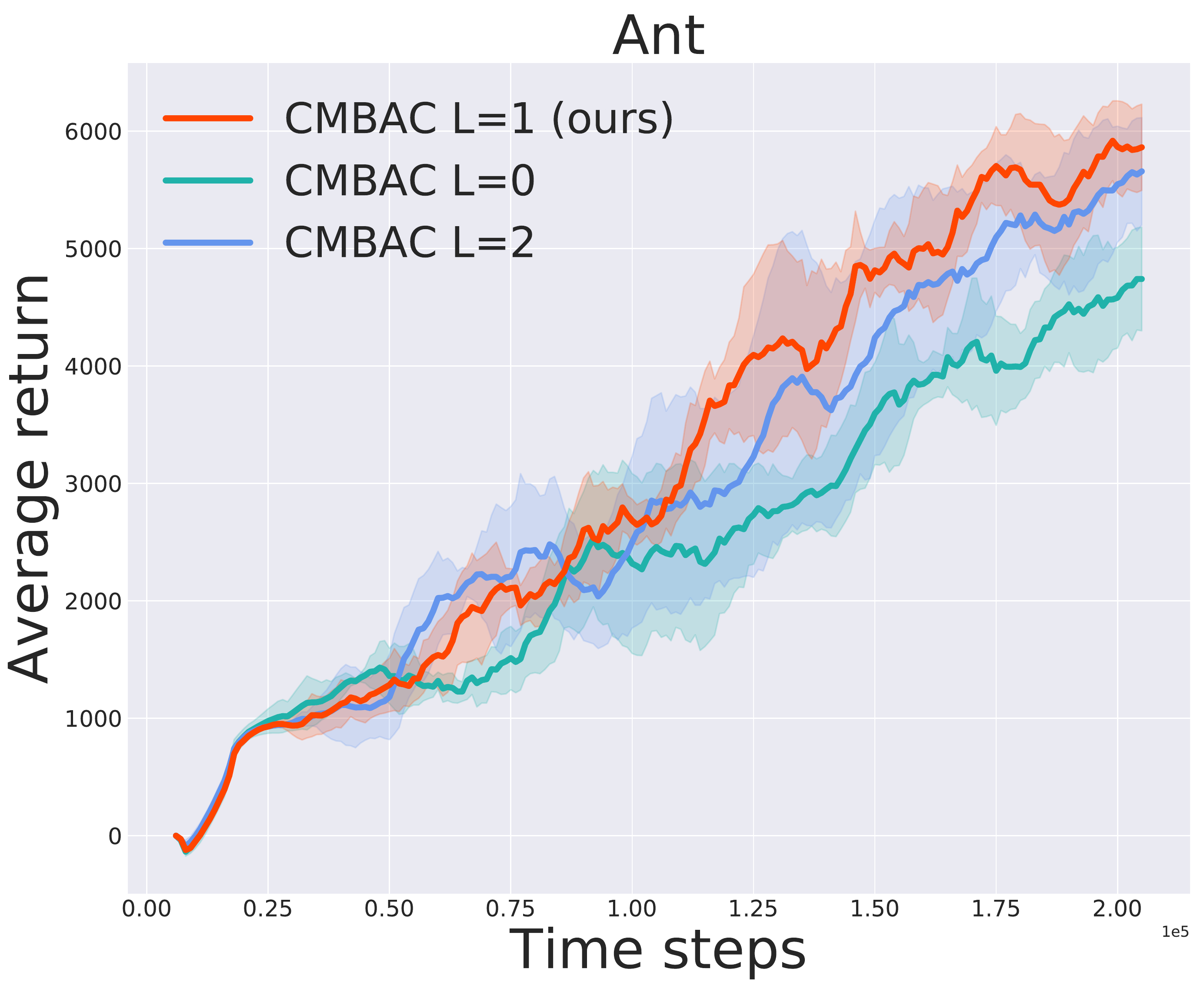}
    \includegraphics[width=0.48\linewidth]{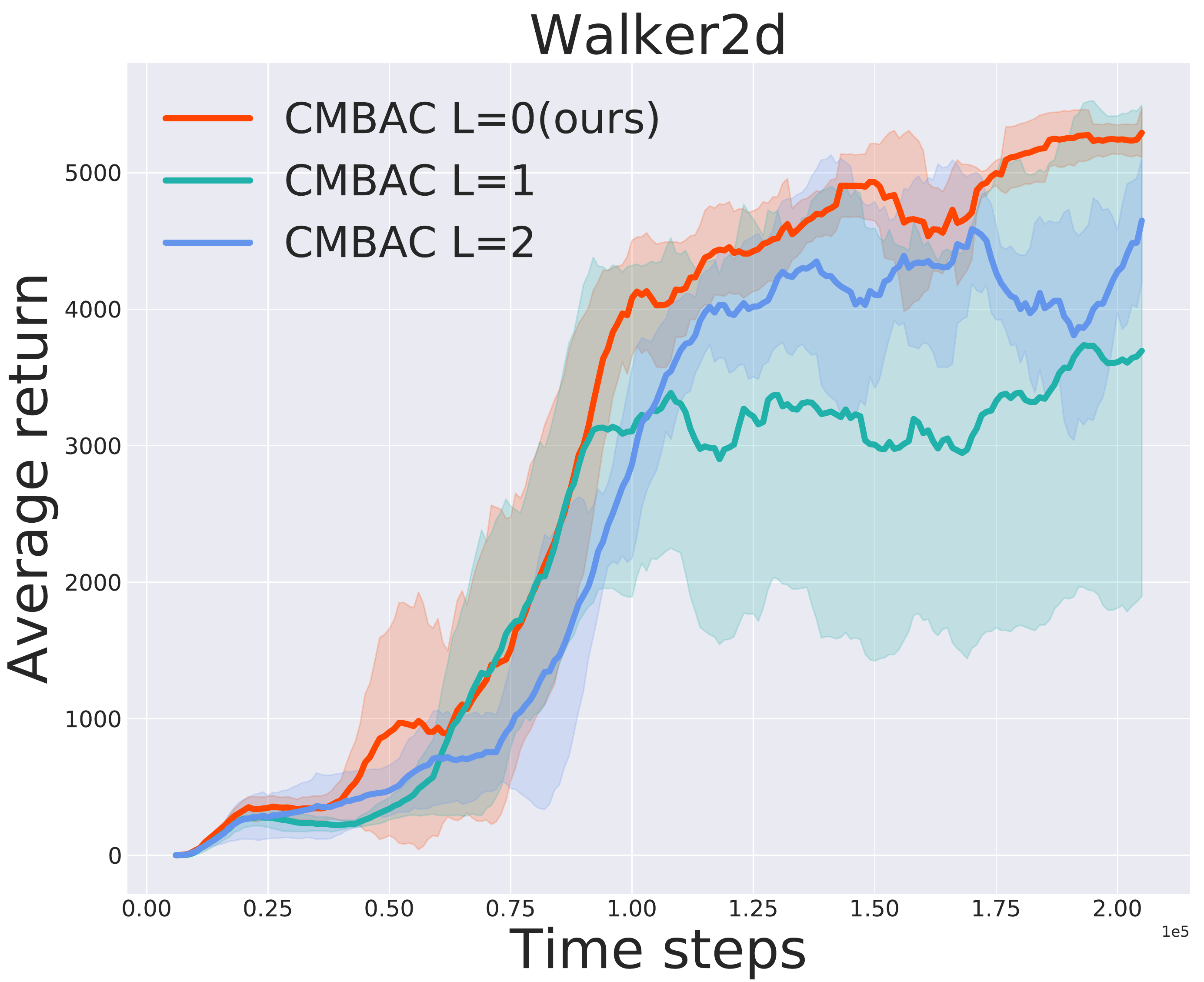}
    \caption{Varying the number of dropped estimates $L$. The results show that $L$ is essential, and $L=1$ provides significant performance improvement on Ant.}
    \label{fig:ablation_drop_num}
\end{minipage}
\end{figure*}

\begin{figure}[t] 
    \centering
    \includegraphics[width=0.48\textwidth]{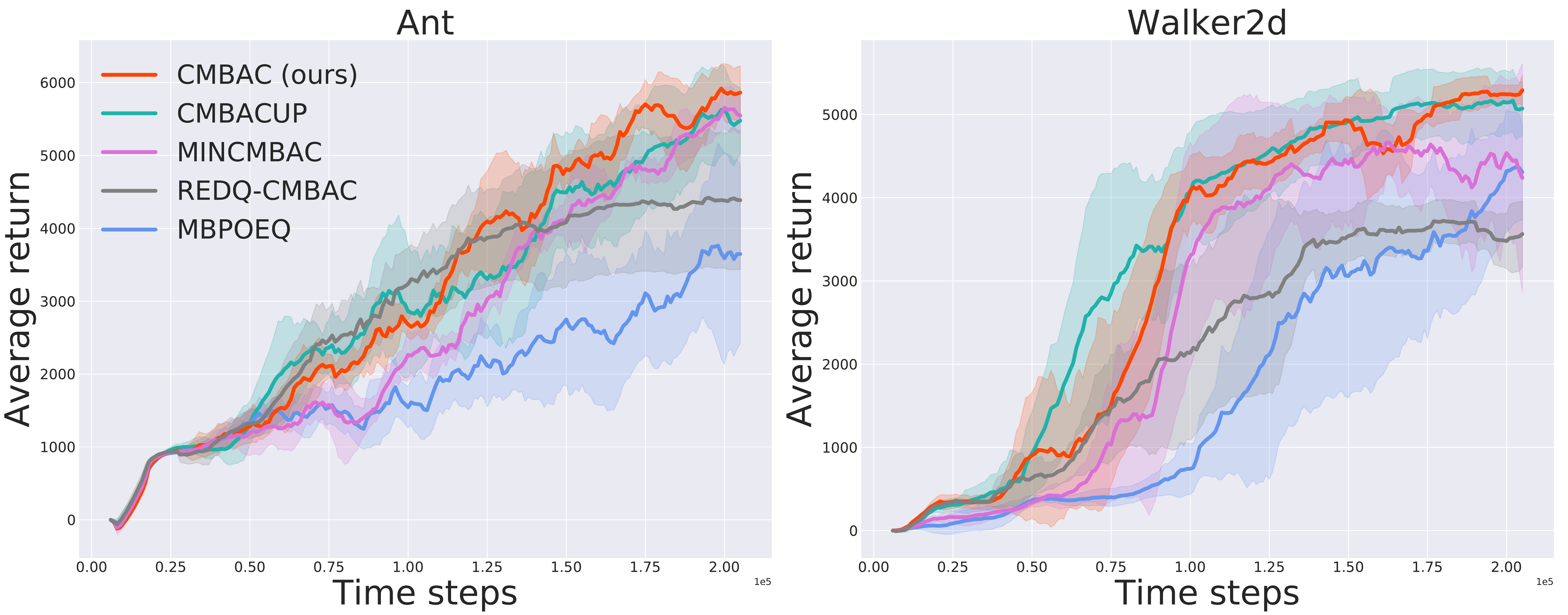}
    \caption{Learning curves of CMBAC and its variants on the Ant and Walker2d environments. The results show that increasing ensemble Q-values does not boost the performance, and different implementations of conservative policy optimization significantly impact performance.}
    \label{fig:CMBAC variants}
\end{figure}

\subsection{Ablation Study}
    In this subsection, we perform carefully designed ablation study to understand the superiority of CMBAC on Ant, which is one of the most challenging environments, and Walker2d, which has the highest resolution power among all 2D environments.  We provide additional results in Appendix C. All results are reported over at least 4 random seeds. 

\subsubsection{Contribution of each component}

    The path from MBPO \cite{janner2019mbpo} to CMBAC comprises three modifications: Q-network size increase (Big), learning multiple estimates of the Q-value (LMEQ), and conservative policy optimization (CPO). B-MBPO is MBPO with an increased size of Q-networks (Big MBPO). The big Q-network has three layers with 512 neurons versus two layers of 256 neurons in MBPO. Note that the policy network size does not change. B-LMEQ-MBPO is B-MBPO with learning multiple estimates of the Q-value to capture the uncertainty of the Q-value function. B-LMEQ-MBPO uses the average of all the estimates to optimize the policy. CMBAC is B-LMEQ-MBPO with conservative policy optimization, i.e., it uses the average of the bottom-k estimates to optimize the policy.  

    Figure \ref{fig:ablation_component} shows that both LMEQ and CPO are significant for performance improvement (though CPO does not improve the performance on Walker2d). The increased Q-network size does not improve MBPO. The uncertainty estimation method of CMBAC improves its performance in both environments. A possible reason is that it may incorporate more information of the model into the multi-head Q-network and thus improves the value estimation. For the Ant environment, conservative policy optimization provides an additional performance improvement. For the Walker2d environment, we find that our method without conservative policy optimization performs better than that with it (please refer to Figure \ref{fig:ablation_drop_num}). One possible reason is that the agent may require an optimistic value estimate to promote exploration, and thus avoid suboptimal policies on Walker2d. 

\subsubsection{Sensitivity to hyperparameters}
    In this part, we analyze the sensitivity of CMBAC to the number of neural networks in each model $M$ and the number of dropped estimates $L$. Please refer to Appendix C for additional results.
    
    \textbf{The number of neural networks in a model $M$} We vary the number $M\in \{1,2,3,4\}$ with $N=5$ on the Ant and Walker2d environments. Figure \ref{fig:ablation_subset_size} shows that (1) the number $M$ is essential as it controls the granularity of uncertainty capturing, and (2) there is an optimal number, i.e., $M=2$. 
    
    \textbf{The number of dropped estimates $L$} We vary the number of dropped estimates for each ensemble multi-head Q-network $L\in\{0,1,2\}$. The total number of estimates dropped is $2L$.
    Figure \ref{fig:ablation_drop_num} suggests that dropping $L$ topmost estimates provides a performance improvement on the Ant environment but degrades the performance on the Walker2d. 

\subsubsection{Analysis of CMBAC variants}
    To provide further insight into CMBAC, we discuss the performance of some CMBAC variants. These variants implement the two core components of CMBAC, i.e., capturing the uncertainty of the Q-value and conservative policy optimization, in different ways. 

    \textbf{LMEQ variants} To capture the uncertainty of the Q-value, we propose to increase the number of ensemble Q-value functions, i.e., deep ensembles \cite{deep_ensemble}. Thus, we propose Model-Based Policy Optimization with Ensemble Q (MBPOEQ), which uses the same multi-head Q-networks as our method. In contrast to CMBAC, MBPOEQ learns all ``heads'' using the same target, i.e., the average of these ``heads''. 
    
    \textbf{CPO variants} We additionally propose three possible conservative policy optimization variants, named CMBAC Uncertainty Penalty (CMBACUP), MINimum CMBAC (MINCMBAC), and REDQ-CMBAC, respectively. CMBACUP uses the standard deviation of the Q-value estimates as a penalty. MINCMBAC uses the minimum of all estimates to optimize the policy. Similar to REDQ \cite{redq}, REDQ-CMBAC
    produces a conservative estimate via each multi-head Q-network and uses the mean of the two conservative estimates to optimize the policy. Due to limited space, we provide the details in Appendix D.
    
    The results in Figure \ref{fig:CMBAC variants} suggest the following conclusions. CMBACUP achieves comparable performance to CMBAC in both environments. However, we find it sensitive to the penalty coefficient (please refer to Appendix C). Increasing the number of ensemble Q-values does not improve performance, indicating that our uncertainty capturing technique is critical to performance improvement. Incorporating conservatism is significant for performance improvement, and different implementations of conservative policy optimization have a large impact on performance.

\begin{figure}[t] 
    \centering
    \includegraphics[width=100pt]{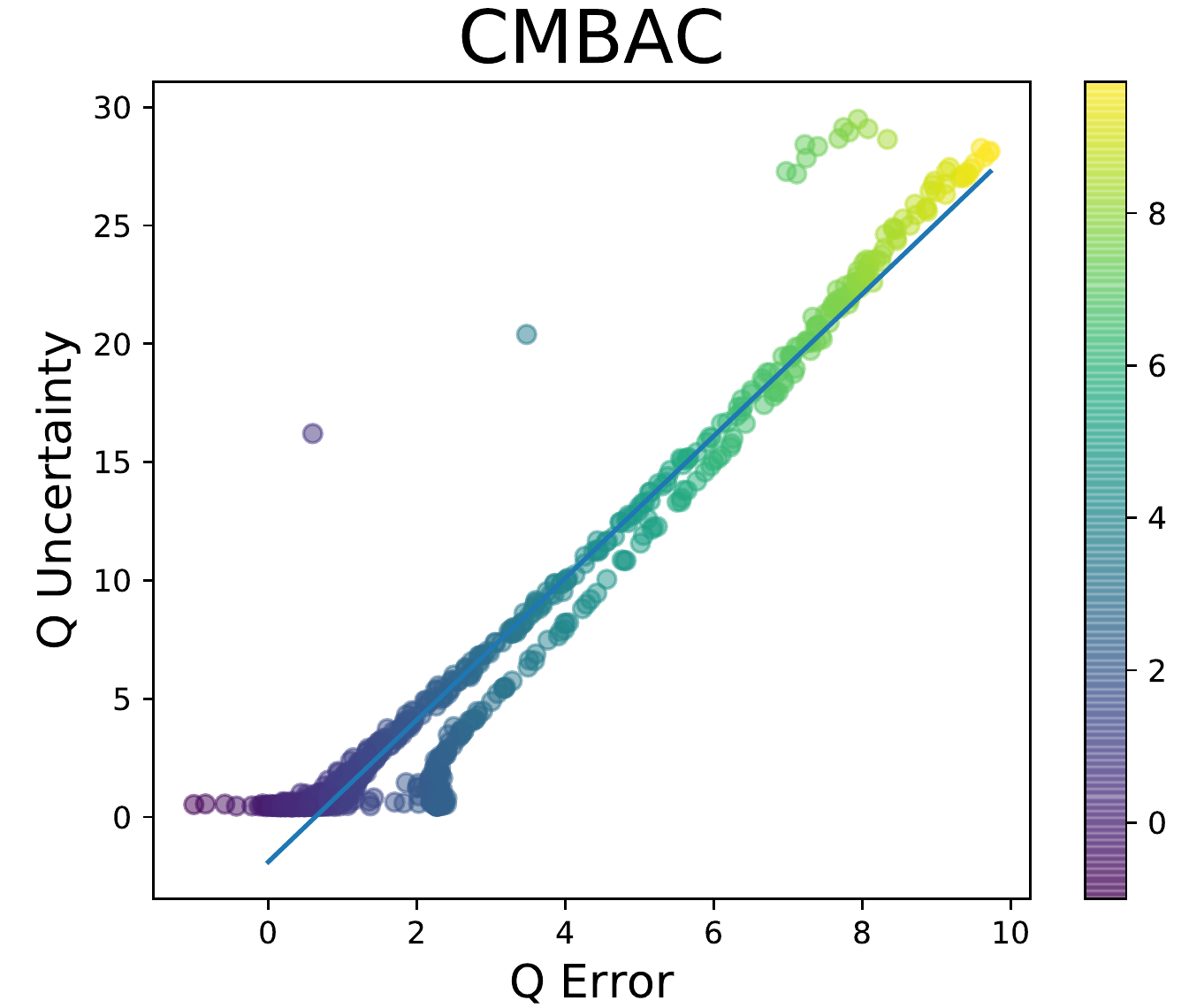}
    \includegraphics[width=100pt]{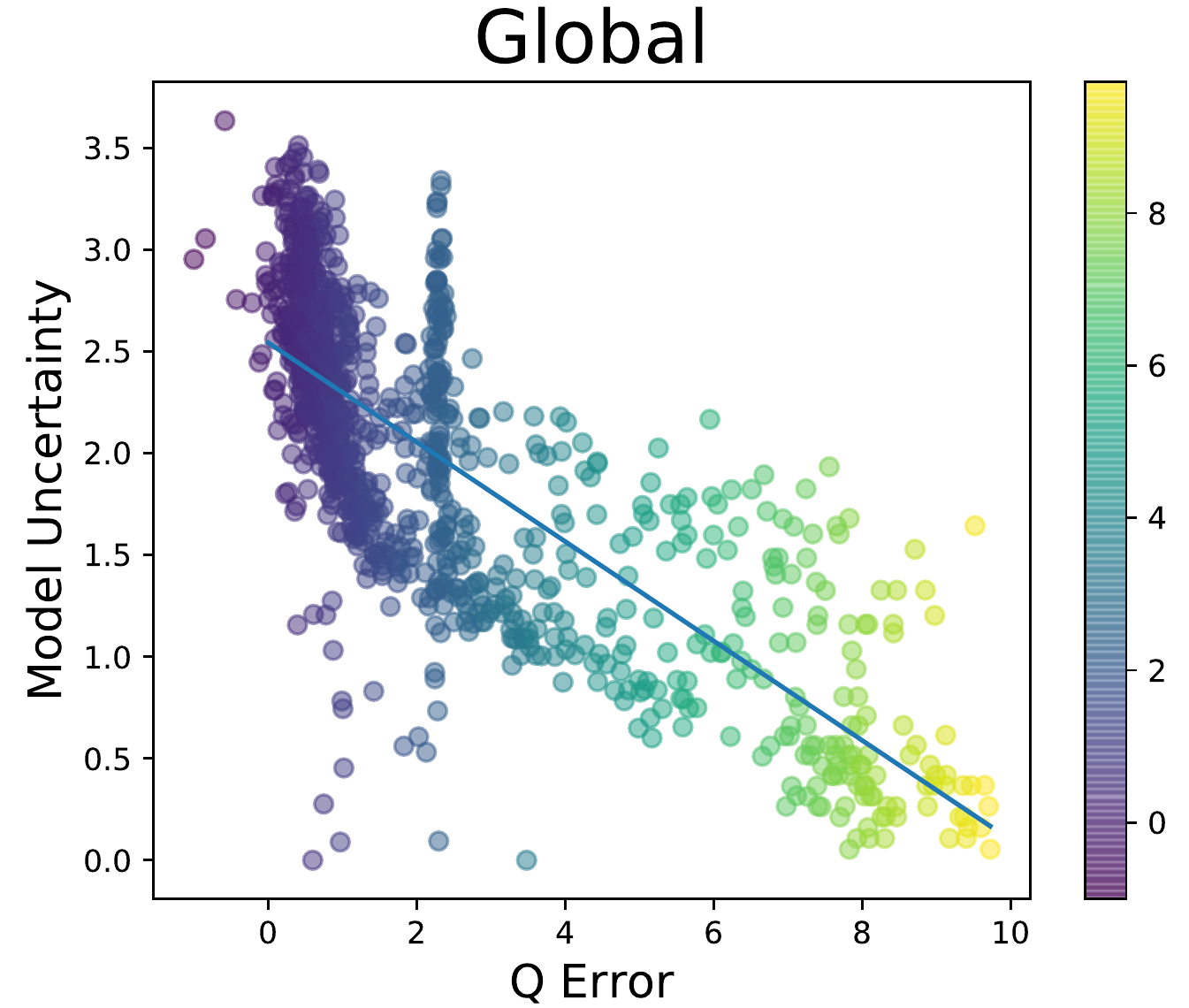}
    \includegraphics[width=100pt]{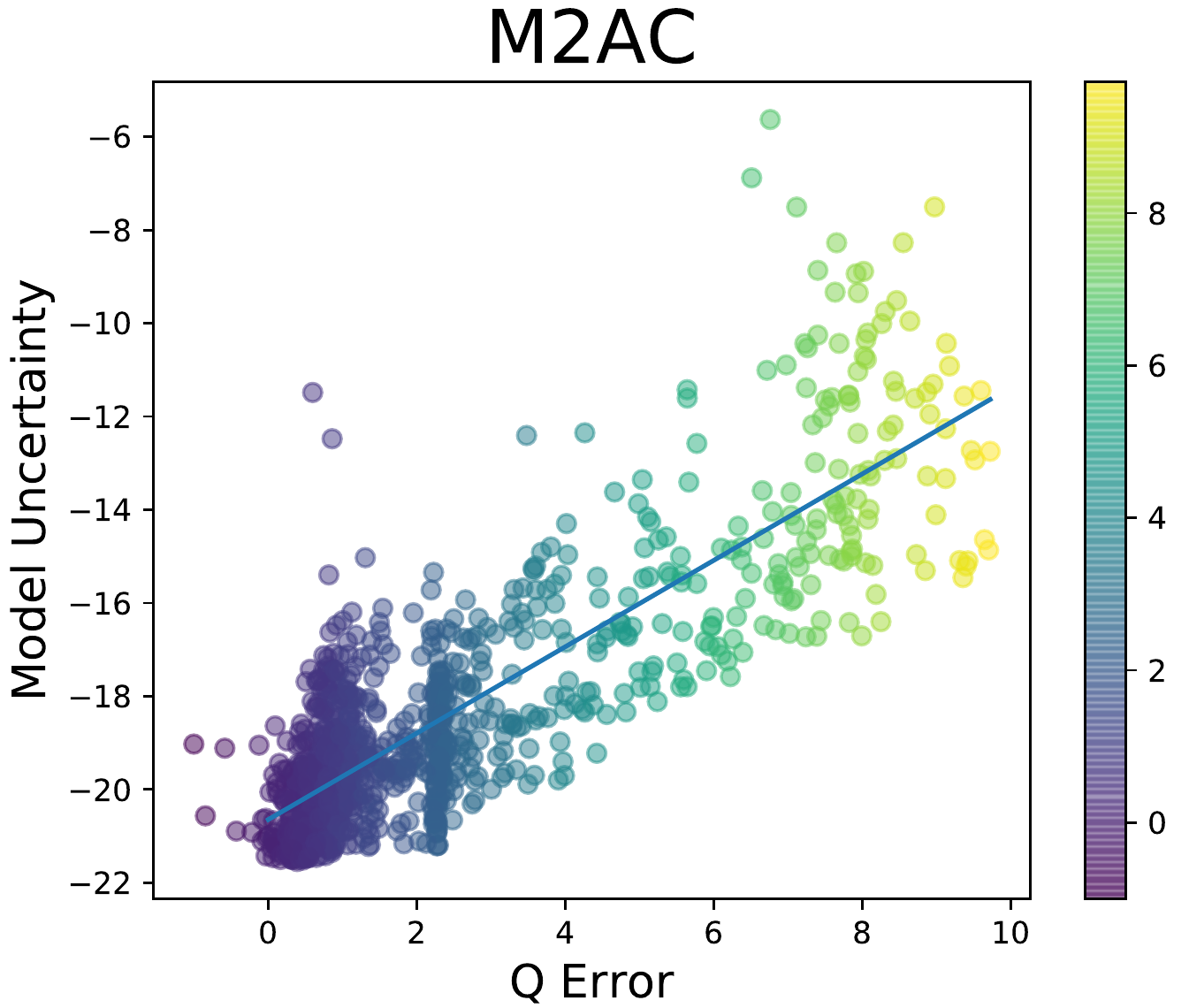}
    \includegraphics[width=100pt]{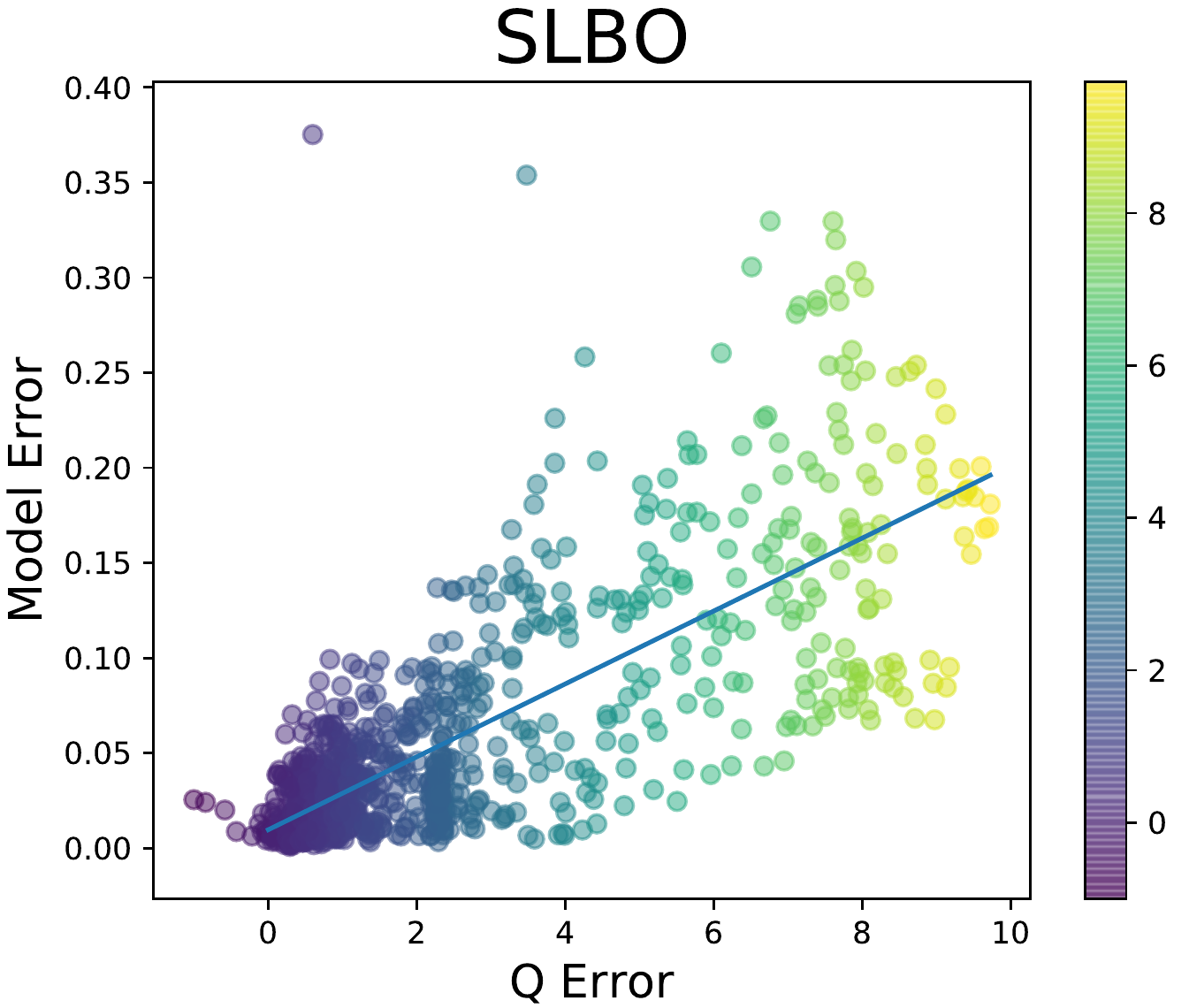}
    \caption{We visualize our uncertainty estimation compared to previous methods on the InvertedPendulum environment. Colored points illustrate the uncertainty versus the value estimation error at given state-action pairs. Results show that our uncertainty estimation approximates the errors of the Q-function more accurately than previous methods. (For a fair comparison, we use a fixed policy to generate rollouts in the true
    environment for computing the real return and use its corresponding Q-value network to estimate the Q-value.) }
\label{fig:visulization}
\end{figure}

\subsection{Visualization}\label{sec:visualization}
    \subsubsection{Uncertainty estimation} 
    In this part, we regard the standard deviation of the multiple heads as our estimated uncertainty and compare it to previous uncertainty estimation used in state-of-the art model-based algorithms. The uncertainty estimation methods include Global, i.e., the cumulative discounted sum of prediction errors similar to that used in model-based offline policy optimization \cite{mopo}, M2AC that is used in masked model-based actor-critic \cite{m2ac}, and SLBO that is used in stochastic lower bounds optimization \cite{slbo}. We visualize the prediction error of the Q-value and the estimated uncertainty via scatters. The results in Figure \ref{fig:visulization} show that our estimated uncertainty can approximate the errors of the Q-function more accurately than previous methods. The superiority of our uncertainty estimation demonstrates that the multi-head Q-networks used in CMBAC can provide a reasonable approximation of the posterior distribution of the true Q-value. Moreover, we find that the cumulative discounted sum of prediction errors can hardly approximate the errors of the Q-function. We provide possible reasons in Appendix D.

    \subsubsection{Conservative policy optimization} 
\begin{figure}[t]
    \centering
    \includegraphics[width=0.18\textwidth]{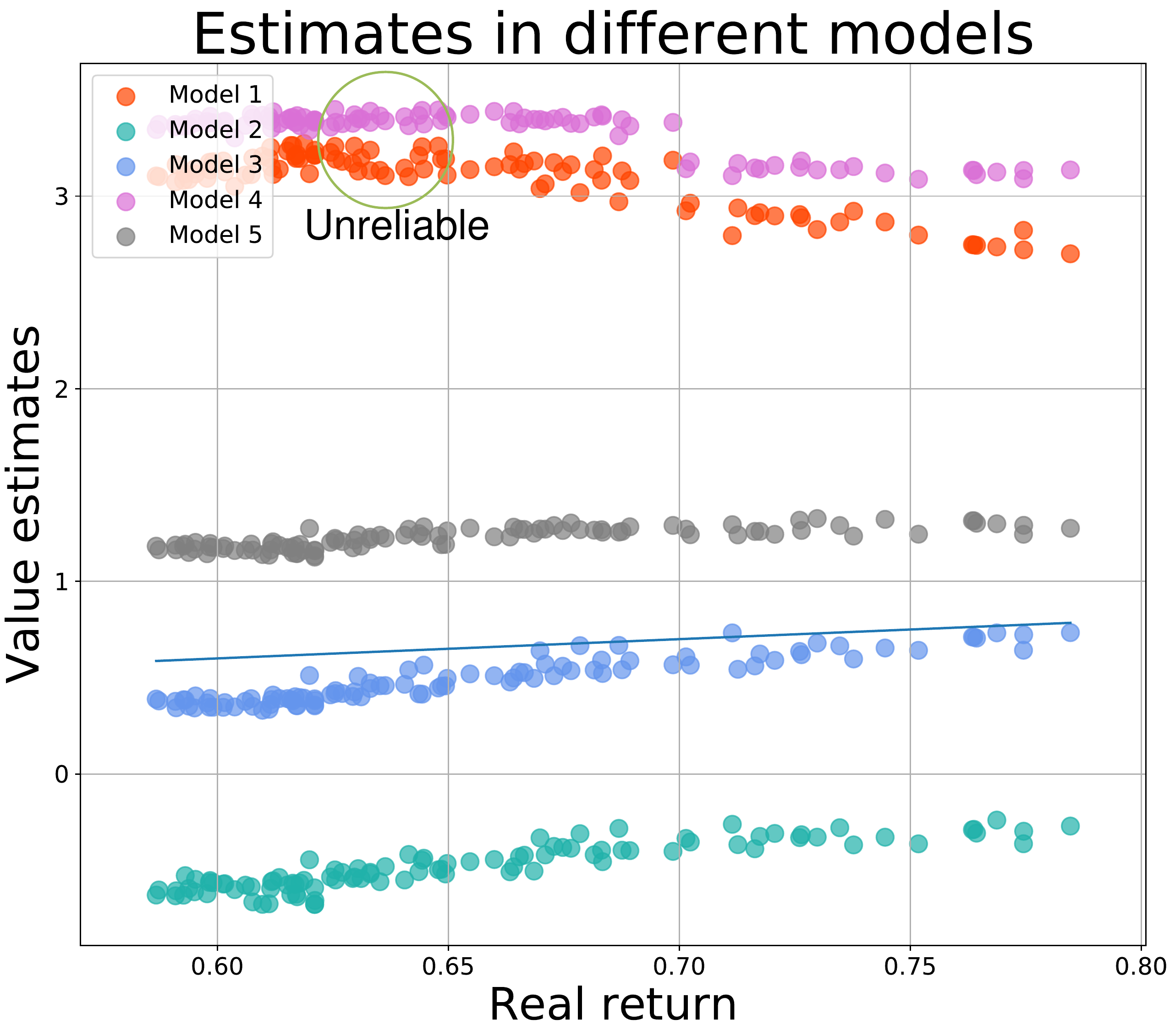}
    \includegraphics[width=0.18\textwidth]{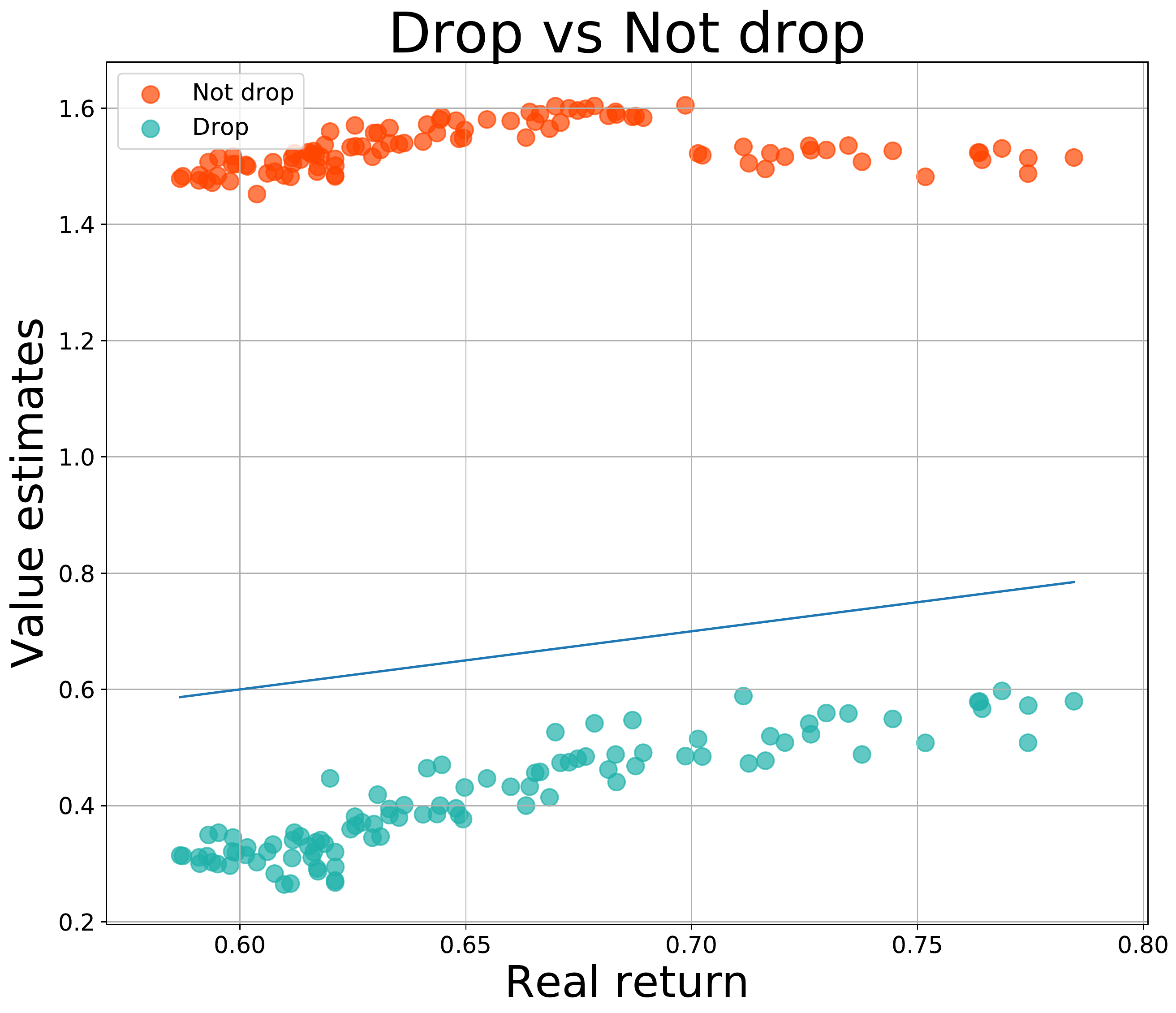}
    \caption{We visualize the estimates of the Q-value in each model on the 2D point environment. Colored points illustrate the estimate versus the true return at given state-action pairs. The results demonstrate that dropping several topmost estimates effectively avoid unreliable actions.
    }
    \label{fig:visulization_model_ret}
\end{figure}

We visualize the estimates of the Q-value in each model versus the true return on the 2D point environment. To better understand the superiority of the proposed conservative policy optimization, we use a simplified version of CMBAC (please refer to Appendix D for details). Figure \ref{fig:visulization_model_ret} (left) shows that there are a small fraction of models in which the value estimates significantly overestimate due to model errors. Moreover, Figure \ref{fig:visulization_model_ret} shows that dropping several topmost estimates effectively encourages the agent to avoid the unreliable ``promising actions''---whose values are high in only a small fraction of models.

\begin{figure}[t]
    \centering
    \includegraphics[height=0.18\textwidth]{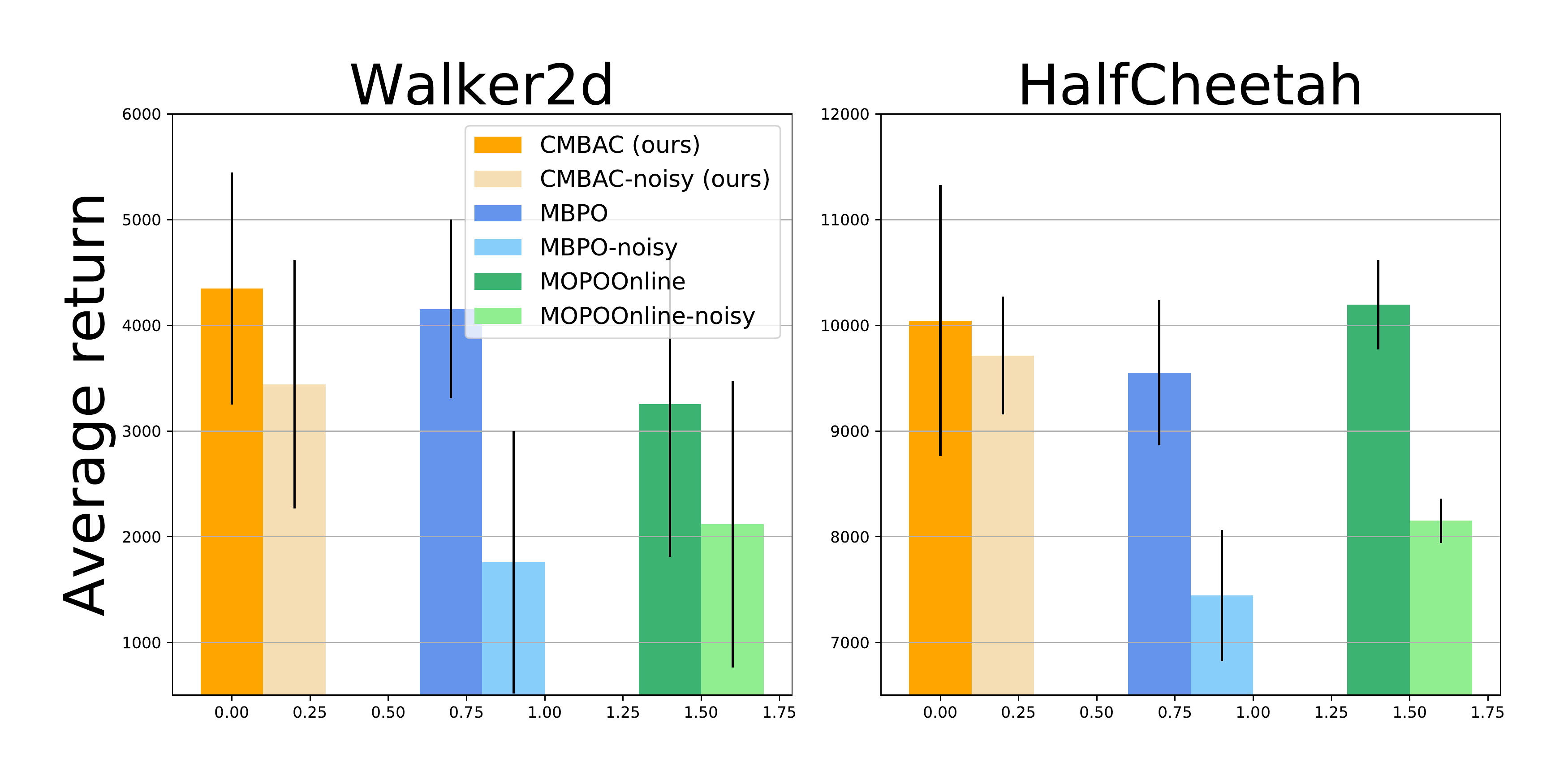}
    \caption{Performance of CMBAC, MBPO, and MOPO-Online in noisy environments (100k steps for HalfCheetah and 200k steps for Walker2d). The results show that our method is more robust than state-of-the-art methods.}
    \label{fig:robust_analysis}
\end{figure}

\subsection{Robust Analysis}
    To understand the robustness of CMBAC, we compare it to MOPO-Online and MBPO on noisy Walker2d and HalfCheetah environments. In these noisy environments, we add Gaussian white noises with the standard deviation $\sigma=0.1$ to the agent's action at every step. The added noise will decrease the accuracy of the learned models. The results in Figure \ref{fig:robust_analysis} show that CMBAC performs robustly and significantly outperforms the baselines in terms of sample efficiency in noisy environments.

\section{Conclusion}
    In this paper, we present \textbf{c}onservative \textbf{m}odel-\textbf{b}ased \textbf{a}ctor-\textbf{c}ritic, a novel approach that approximates a posterior distribution over Q-values based on the ensemble models and uses the average of the left tail of the distribution approximation to optimize the policy.
    Experiments show that CMBAC significantly outperforms state-of-the-art methods in terms of sample efficiency on several challenging control tasks. Moreover, experiments demonstrate that the proposed method is more robust than previous methods in noisy environments. 
    We believe that our proposed approach will bring new insights into model-based reinforcement learning. 

\section*{Acknowledgements}

We would like to thank all the anonymous reviewers for their insightful comments. This work was supported in part by National Science Foundations of China grants U19B2026, U19B2044, 61822604, 61836006, and 62021001, and the Fundamental Research Funds for the Central Universities grant WK3490000004.

\bibliography{aaai22}

\begin{thebibliography}{43}
\providecommand{\natexlab}[1]{#1}

\bibitem[{Amos et~al.(2021)Amos, Stanton, Yarats, and Wilson}]{sac_svg}
Amos, B.; Stanton, S.; Yarats, D.; and Wilson, A.~G. 2021.
\newblock On the Model-Based Stochastic Value Gradient for Continuous
  Reinforcement Learning learning.
\newblock In \emph{Proceedings of the 3rd Conference on Learning for Dynamics
  and Control}, volume 144 of \emph{Proceedings of Machine Learning Research},
  6--20. PMLR.

\bibitem[{Brockman et~al.(2016)Brockman, Cheung, Pettersson, Schneider,
  Schulman, Tang, and Zaremba}]{openaigym}
Brockman, G.; Cheung, V.; Pettersson, L.; Schneider, J.; Schulman, J.; Tang,
  J.; and Zaremba, W. 2016.
\newblock OpenAI Gym.
\newblock \emph{CoRR}, abs/1606.01540.

\bibitem[{Chen et~al.(2021)Chen, Wang, Zhou, and Ross}]{redq}
Chen, X.; Wang, C.; Zhou, Z.; and Ross, K.~W. 2021.
\newblock Randomized Ensembled Double Q-Learning: Learning Fast Without a
  Model.
\newblock In \emph{International Conference on Learning Representations}.

\bibitem[{Chua et~al.(2018)Chua, Calandra, McAllister, and
  Levine}]{chua2018pets}
Chua, K.; Calandra, R.; McAllister, R.; and Levine, S. 2018.
\newblock Deep Reinforcement Learning in a Handful of Trials using
  Probabilistic Dynamics Models.
\newblock In \emph{Advances in Neural Information Processing Systems},
  volume~31.

\bibitem[{Clavera, Fu, and Abbeel(2020)}]{maac}
Clavera, I.; Fu, Y.; and Abbeel, P. 2020.
\newblock Model-Augmented Actor-Critic: Backpropagating through Paths.
\newblock In \emph{International Conference on Learning Representations}.

\bibitem[{Clavera et~al.(2018)Clavera, Rothfuss, Schulman, Fujita, Asfour, and
  Abbeel}]{mb_mpo}
Clavera, I.; Rothfuss, J.; Schulman, J.; Fujita, Y.; Asfour, T.; and Abbeel, P.
  2018.
\newblock Model-Based Reinforcement Learning via Meta-Policy Optimization.
\newblock In \emph{Proceedings of The 2nd Conference on Robot Learning},
  volume~87 of \emph{Proceedings of Machine Learning Research}, 617--629. PMLR.

\bibitem[{de~Boer et~al.(2005)de~Boer, Kroese, Mannor, and Rubinstein}]{cem}
de~Boer, P.; Kroese, D.~P.; Mannor, S.; and Rubinstein, R.~Y. 2005.
\newblock A Tutorial on the Cross-Entropy Method.
\newblock \emph{Ann. Oper. Res.}, 134(1): 19--67.

\bibitem[{Deisenroth and Rasmussen(2011)}]{deisenroth2011pilco}
Deisenroth, M.; and Rasmussen, C.~E. 2011.
\newblock PILCO: A Model-Based and Data-Efficient Approach to Policy Search.
\newblock In \emph{Proceedings of the 28th International Conference on Machine
  Learning (ICML-11)}, ICML '11, 465--472. ACM.
\newblock ISBN 978-1-4503-0619-5.

\bibitem[{Doyle(1996)}]{robust_optimal_control}
Doyle, J. 1996.
\newblock Robust and optimal control.
\newblock In \emph{Proceedings of 35th IEEE Conference on Decision and
  Control}, volume~2, 1595--1598 vol.2.

\bibitem[{Efron and Tibshirani(1993)}]{introduction_bootstrap}
Efron, B.; and Tibshirani, R. 1993.
\newblock \emph{An Introduction to the Bootstrap}.
\newblock Springer.
\newblock ISBN 978-1-4899-4541-9.

\bibitem[{Fairbank and Alonso(2012)}]{vgl}
Fairbank, M.; and Alonso, E. 2012.
\newblock Value-gradient learning.
\newblock In \emph{The 2012 International Joint Conference on Neural Networks
  (IJCNN)}, 1--8.

\bibitem[{Fujimoto, van Hoof, and Meger(2018)}]{td3}
Fujimoto, S.; van Hoof, H.; and Meger, D. 2018.
\newblock Addressing Function Approximation Error in Actor-Critic Methods.
\newblock In \emph{Proceedings of the 35th International Conference on Machine
  Learning}, volume~80 of \emph{Proceedings of Machine Learning Research},
  1587--1596. PMLR.

\bibitem[{Haarnoja et~al.(2017)Haarnoja, Tang, Abbeel, and Levine}]{sql}
Haarnoja, T.; Tang, H.; Abbeel, P.; and Levine, S. 2017.
\newblock Reinforcement Learning with Deep Energy-Based Policies.
\newblock In \emph{Proceedings of the 34th International Conference on Machine
  Learning}, volume~70 of \emph{Proceedings of Machine Learning Research},
  1352--1361. PMLR.

\bibitem[{Haarnoja et~al.(2018)Haarnoja, Zhou, Abbeel, and Levine}]{sac}
Haarnoja, T.; Zhou, A.; Abbeel, P.; and Levine, S. 2018.
\newblock Soft Actor-Critic: Off-Policy Maximum Entropy Deep Reinforcement
  Learning with a Stochastic Actor.
\newblock In \emph{Proceedings of the 35th International Conference on Machine
  Learning}, volume~80 of \emph{Proceedings of Machine Learning Research},
  1861--1870. PMLR.

\bibitem[{Heess et~al.(2015)Heess, Wayne, Silver, Lillicrap, Erez, and
  Tassa}]{svg}
Heess, N.; Wayne, G.; Silver, D.; Lillicrap, T.; Erez, T.; and Tassa, Y. 2015.
\newblock Learning Continuous Control Policies by Stochastic Value Gradients.
\newblock In \emph{Advances in Neural Information Processing Systems},
  volume~28.

\bibitem[{Hessel et~al.(2018)Hessel, Modayil, Van~Hasselt, Schaul, Ostrovski,
  Dabney, Horgan, Piot, Azar, and Silver}]{hessel2018rainbow}
Hessel, M.; Modayil, J.; Van~Hasselt, H.; Schaul, T.; Ostrovski, G.; Dabney,
  W.; Horgan, D.; Piot, B.; Azar, M.; and Silver, D. 2018.
\newblock Rainbow: Combining improvements in deep reinforcement learning.
\newblock In \emph{Thirty-second AAAI conference on artificial intelligence}.

\bibitem[{Janner et~al.(2019)Janner, Fu, Zhang, and Levine}]{janner2019mbpo}
Janner, M.; Fu, J.; Zhang, M.; and Levine, S. 2019.
\newblock When to Trust Your Model: Model-Based Policy Optimization.
\newblock In \emph{Advances in Neural Information Processing Systems},
  volume~32. Curran Associates, Inc.

\bibitem[{Kumar et~al.(2020)Kumar, Zhou, Tucker, and Levine}]{cql}
Kumar, A.; Zhou, A.; Tucker, G.; and Levine, S. 2020.
\newblock Conservative Q-Learning for Offline Reinforcement Learning.
\newblock In \emph{Advances in Neural Information Processing Systems},
  volume~33, 1179--1191. Curran Associates, Inc.

\bibitem[{Kurutach et~al.(2018)Kurutach, Clavera, Duan, Tamar, and
  Abbeel}]{kurutach2018metrpo}
Kurutach, T.; Clavera, I.; Duan, Y.; Tamar, A.; and Abbeel, P. 2018.
\newblock Model-Ensemble Trust-Region Policy Optimization.
\newblock In \emph{International Conference on Learning Representations}.

\bibitem[{Kuznetsov et~al.(2020)Kuznetsov, Shvechikov, Grishin, and
  Vetrov}]{tqc}
Kuznetsov, A.; Shvechikov, P.; Grishin, A.; and Vetrov, D. 2020.
\newblock Controlling Overestimation Bias with Truncated Mixture of Continuous
  Distributional Quantile Critics.
\newblock In \emph{Proceedings of the 37th International Conference on Machine
  Learning}, volume 119 of \emph{Proceedings of Machine Learning Research},
  5556--5566. PMLR.

\bibitem[{Lakshminarayanan, Pritzel, and Blundell(2017)}]{deep_ensemble}
Lakshminarayanan, B.; Pritzel, A.; and Blundell, C. 2017.
\newblock Simple and Scalable Predictive Uncertainty Estimation using Deep
  Ensembles.
\newblock In \emph{Advances in Neural Information Processing Systems},
  volume~30. Curran Associates, Inc.

\bibitem[{Lillicrap et~al.(2016)Lillicrap, Hunt, Pritzel, Heess, Erez, Tassa,
  Silver, and Wierstra}]{lillicrap2015continuous}
Lillicrap, T.~P.; Hunt, J.~J.; Pritzel, A.; Heess, N.; Erez, T.; Tassa, Y.;
  Silver, D.; and Wierstra, D. 2016.
\newblock Continuous control with deep reinforcement learning.
\newblock In \emph{4th International Conference on Learning Representations,
  {ICLR} 2016}.

\bibitem[{Lim, Xu, and Mannor(2013)}]{NIPS2013_robust_mdp}
Lim, S.~H.; Xu, H.; and Mannor, S. 2013.
\newblock Reinforcement Learning in Robust Markov Decision Processes.
\newblock In \emph{Advances in Neural Information Processing Systems},
  volume~26. Curran Associates, Inc.

\bibitem[{Luo et~al.(2019)Luo, Xu, Li, Tian, Darrell, and Ma}]{slbo}
Luo, Y.; Xu, H.; Li, Y.; Tian, Y.; Darrell, T.; and Ma, T. 2019.
\newblock Algorithmic Framework for Model-based Deep Reinforcement Learning
  with Theoretical Guarantees.
\newblock In \emph{International Conference on Learning Representations}.

\bibitem[{Mnih et~al.(2015)Mnih, Kavukcuoglu, Silver, Rusu, Veness, Bellemare,
  Graves, Riedmiller, Fidjeland, Ostrovski et~al.}]{nature_dqn}
Mnih, V.; Kavukcuoglu, K.; Silver, D.; Rusu, A.~A.; Veness, J.; Bellemare,
  M.~G.; Graves, A.; Riedmiller, M.; Fidjeland, A.~K.; Ostrovski, G.; et~al.
  2015.
\newblock Human-level control through deep reinforcement learning.
\newblock \emph{nature}, 518(7540): 529--533.

\bibitem[{Nagabandi et~al.(2018)Nagabandi, Kahn, Fearing, and
  Levine}]{NagabandiKFL18}
Nagabandi, A.; Kahn, G.; Fearing, R.~S.; and Levine, S. 2018.
\newblock Neural Network Dynamics for Model-Based Deep Reinforcement Learning
  with Model-Free Fine-Tuning.
\newblock In \emph{2018 IEEE International Conference on Robotics and
  Automation (ICRA)}, 7559--7566.

\bibitem[{Nguyen and Widrow(1990)}]{nguyen1990neural}
Nguyen, D.; and Widrow, B. 1990.
\newblock Neural networks for self-learning control systems.
\newblock \emph{IEEE Control Systems Magazine}, 10(3): 18--23.

\bibitem[{O'Donoghue et~al.(2018)O'Donoghue, Osband, Munos, and Mnih}]{ube}
O'Donoghue, B.; Osband, I.; Munos, R.; and Mnih, V. 2018.
\newblock The Uncertainty {B}ellman Equation and Exploration.
\newblock In \emph{Proceedings of the 35th International Conference on Machine
  Learning}, volume~80 of \emph{Proceedings of Machine Learning Research},
  3839--3848. PMLR.

\bibitem[{Osband et~al.(2016)Osband, Blundell, Pritzel, and
  Van~Roy}]{bootstrap_dqn}
Osband, I.; Blundell, C.; Pritzel, A.; and Van~Roy, B. 2016.
\newblock Deep Exploration via Bootstrapped DQN.
\newblock In \emph{Advances in Neural Information Processing Systems},
  volume~29. Curran Associates, Inc.

\bibitem[{Ovadia et~al.(2019)Ovadia, Fertig, Ren, Nado, Sculley, Nowozin,
  Dillon, Lakshminarayanan, and Snoek}]{trust_uncertainty}
Ovadia, Y.; Fertig, E.; Ren, J.; Nado, Z.; Sculley, D.; Nowozin, S.; Dillon,
  J.; Lakshminarayanan, B.; and Snoek, J. 2019.
\newblock Can you trust your model\textquotesingle s uncertainty? Evaluating
  predictive uncertainty under dataset shift.
\newblock In \emph{Advances in Neural Information Processing Systems},
  volume~32.

\bibitem[{Pan et~al.(2020)Pan, He, Tu, and He}]{m2ac}
Pan, F.; He, J.; Tu, D.; and He, Q. 2020.
\newblock Trust the Model When It Is Confident: Masked Model-based
  Actor-Critic.
\newblock In \emph{Advances in Neural Information Processing Systems},
  volume~33, 10537--10546.

\bibitem[{Punjani and Abbeel(2015)}]{PunjaniA15}
Punjani, A.; and Abbeel, P. 2015.
\newblock Deep learning helicopter dynamics models.
\newblock In \emph{2015 IEEE International Conference on Robotics and
  Automation (ICRA)}, 3223--3230.

\bibitem[{Rajeswaran et~al.(2017)Rajeswaran, Ghotra, Ravindran, and
  Levine}]{epopt}
Rajeswaran, A.; Ghotra, S.; Ravindran, B.; and Levine, S. 2017.
\newblock EPOpt: Learning Robust Neural Network Policies Using Model Ensembles.
\newblock In \emph{5th International Conference on Learning Representations,
  {ICLR} 2017, Toulon, France, April 24-26, 2017, Conference Track
  Proceedings}. OpenReview.net.

\bibitem[{Strehl and Littman(2008)}]{mbieeb}
Strehl, A.~L.; and Littman, M.~L. 2008.
\newblock An analysis of model-based Interval Estimation for Markov Decision
  Processes.
\newblock \emph{J. Comput. Syst. Sci.}, 74(8): 1309--1331.

\bibitem[{Sutton(1990)}]{sutton_dyna_style}
Sutton, R.~S. 1990.
\newblock Integrated Architectures for Learning, Planning, and Reacting Based
  on Approximating Dynamic Programming.
\newblock In \emph{Machine Learning, Proceedings of the Seventh International
  Conference on Machine Learning, Austin, Texas, USA, June 21-23, 1990},
  216--224. Morgan Kaufmann.

\bibitem[{Sutton and Barto(2018)}]{sutton2018reinforcement}
Sutton, R.~S.; and Barto, A.~G. 2018.
\newblock \emph{Reinforcement learning: An introduction}.
\newblock MIT press.

\bibitem[{Todorov, Erez, and Tassa(2012)}]{mujoco}
Todorov, E.; Erez, T.; and Tassa, Y. 2012.
\newblock MuJoCo: A physics engine for model-based control.
\newblock In \emph{2012 IEEE/RSJ International Conference on Intelligent Robots
  and Systems}, 5026--5033.

\bibitem[{Vuong and Tran(2019)}]{uambpo}
Vuong, T.; and Tran, K. 2019.
\newblock Uncertainty-aware Model-based Policy Optimization.
\newblock \emph{CoRR}, abs/1906.10717.

\bibitem[{Wang and Ba(2020)}]{planning_with_net}
Wang, T.; and Ba, J. 2020.
\newblock Exploring Model-based Planning with Policy Networks.
\newblock In \emph{International Conference on Learning Representations}.

\bibitem[{Wang et~al.(2019)Wang, Bao, Clavera, Hoang, Wen, Langlois, Zhang,
  Zhang, Abbeel, and Ba}]{benchmark}
Wang, T.; Bao, X.; Clavera, I.; Hoang, J.; Wen, Y.; Langlois, E.; Zhang, S.;
  Zhang, G.; Abbeel, P.; and Ba, J. 2019.
\newblock Benchmarking Model-Based Reinforcement Learning.
\newblock \emph{CoRR}, abs/1907.02057.

\bibitem[{Yu et~al.(2021)Yu, Kumar, Rafailov, Rajeswaran, Levine, and
  Finn}]{combo}
Yu, T.; Kumar, A.; Rafailov, R.; Rajeswaran, A.; Levine, S.; and Finn, C. 2021.
\newblock {COMBO:} Conservative Offline Model-Based Policy Optimization.
\newblock \emph{CoRR}, abs/2102.08363.

\bibitem[{Yu et~al.(2020)Yu, Thomas, Yu, Ermon, Zou, Levine, Finn, and
  Ma}]{mopo}
Yu, T.; Thomas, G.; Yu, L.; Ermon, S.; Zou, J.~Y.; Levine, S.; Finn, C.; and
  Ma, T. 2020.
\newblock MOPO: Model-based Offline Policy Optimization.
\newblock In \emph{Advances in Neural Information Processing Systems},
  volume~33, 14129--14142. Curran Associates, Inc.

\bibitem[{Zhou, Li, and Wang(2020)}]{pombu}
Zhou, Q.; Li, H.; and Wang, J. 2020.
\newblock Deep Model-Based Reinforcement Learning via Estimated Uncertainty and
  Conservative Policy Optimization.
\newblock \emph{Proceedings of the AAAI Conference on Artificial Intelligence},
  34(04): 6941--6948.

\end{thebibliography}

\newpage

\appendix

\section{Algorithm Implementation Details}
    \subsection{Details of models}
    To enhance the reproducibility of our method, we provide additional details of model learning and model usage. As mentioned in Section 4.1, we represent the model as a set of probabilistic neural networks. The probabilistic neural network (NN) outputs a Gaussian distribution over the next state and reward given the current state and action. Following model-based policy optimization (MBPO) \cite{janner2019mbpo}, we train an ensemble of 7 such probabilistic neural networks and pick the best 5 networks based on the validation prediction error on a held-out set.
    
    We represent the model as a set of $M$ probabilistic neural networks with $M < 5$. Note that $M$ is a fixed hyperparameter in our experiments. We use an ensemble of such models, denoted by $\mathcal{M}$, and the number of models is $K=\tbinom{5}{M}$. For example, if $M=2$, then we use an ensemble of $10$ models. During model rollouts, we randomly pick one dynamics model $\mathcal{M}_j$ from $\mathcal{M}$ and randomly pick one probabilistic neural network from $\mathcal{M}_j$ to generate model dataset $\mathcal{D}_{\text{model}}$, which is the same with MBPO. Moreover, to learn the Q-value estimate in each model, we preserve next states $s_{j}^{\prime}$ generated by each model $\mathcal{M}_j$ given the state $s$ and  action $a$. Since we can generate these next states in parallel, the increased computational cost is low as shown in Figure \ref{fig:wall_clock}. 
    
    \subsection{Difference between CMBAC and robust policy optimization}\label{alg:dis}
        As mentioned in Section 4.2, the conservative policy optimization aims to seek a policy that has high Q-values in most models. Specifically, CMBAC aims to learn the Q-value in every model $\mathcal{M}_j$, denoted by $Q^{\pi, P_j}$. Then given the state $s$ and action $a$, CMBAC uses the average of the bottom-k Q-value estimates pointwisely to optimize the policy. In contrast to robust policy optimization \cite{robust_optimal_control, NIPS2013_robust_mdp, epopt}, the Q-value estimate, i.e., the average of the bottom-k estimates, used in CMBAC may correspond to different models at different state-action pairs. Therefore, the Q-value estimate may not correspond to any individual transition function. Even if CMBAC uses the minimum estimate to optimize the policy, it is still different from robust policy optimization. 
        
        

\section{Experimental Settings and Hyperparameters}
\subsection{Reproducing the baselines}
    We first list the implementations of our baselines, including model-based policy optimization (MBPO) \cite{janner2019mbpo}, soft actor-critic (SAC) \cite{sac}, and randomized ensembled double Q-learning (REDQ) \cite{redq}. We then present the implementation details of MOPO-Online, i.e., the online variant of model-based offline policy optimization (MOPO) \cite{mopo}. 
    
    \subsubsection{MBPO} We use the official data provided by the authors of MBPO in our figures. 
        
    \subsubsection{SAC} We use the PyTorch implementation of SAC in the rlkit repository, which is recommended by the authors of SAC, to reproduce the results. 
    
    \subsubsection{REDQ} We use the official code provided by the authors  of REDQ to reproduce the results. 
    
    \subsubsection{MOPO-Online} We implement MOPO-Online on top of the MBPO. We quantify the uncertainy as proposed in \citet{mopo}, i.e., 
    \begin{align}\label{eq:mopo_uncertainty}
        u(s,a) = \max_{i=1,\dots,7} \|\Sigma_i(s,a)\|_{F},
    \end{align}
    where $\Sigma_i(s,a)$ is the diagonal covariance given by each probabilistic neural network. We then use the uncertainty as a reward penalty
    \begin{align}
        r_p(s,a) = r(s,a) - c \cdot u(s,a),
    \end{align}
    where $c$ is a constant. 
    
\begin{figure*}[t] 
\centering
    \includegraphics[width=0.64\linewidth]{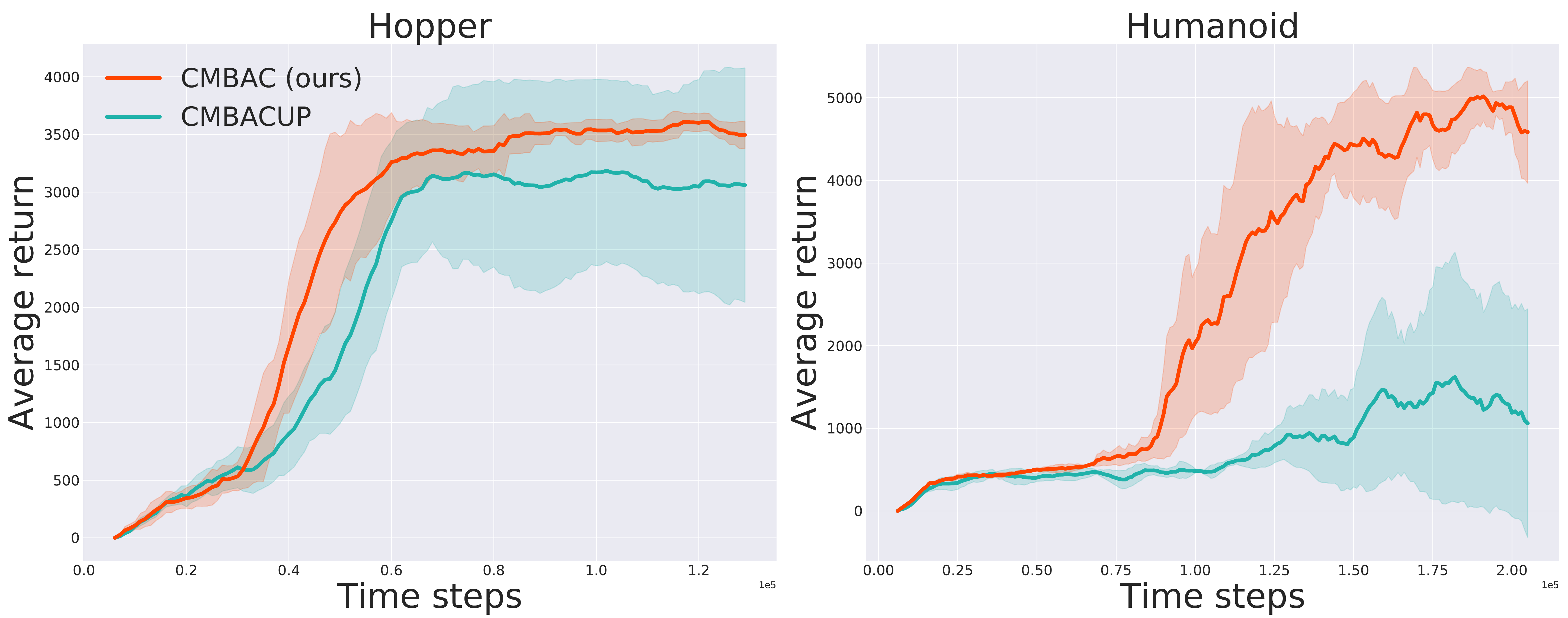}
    \includegraphics[width=0.32\linewidth]{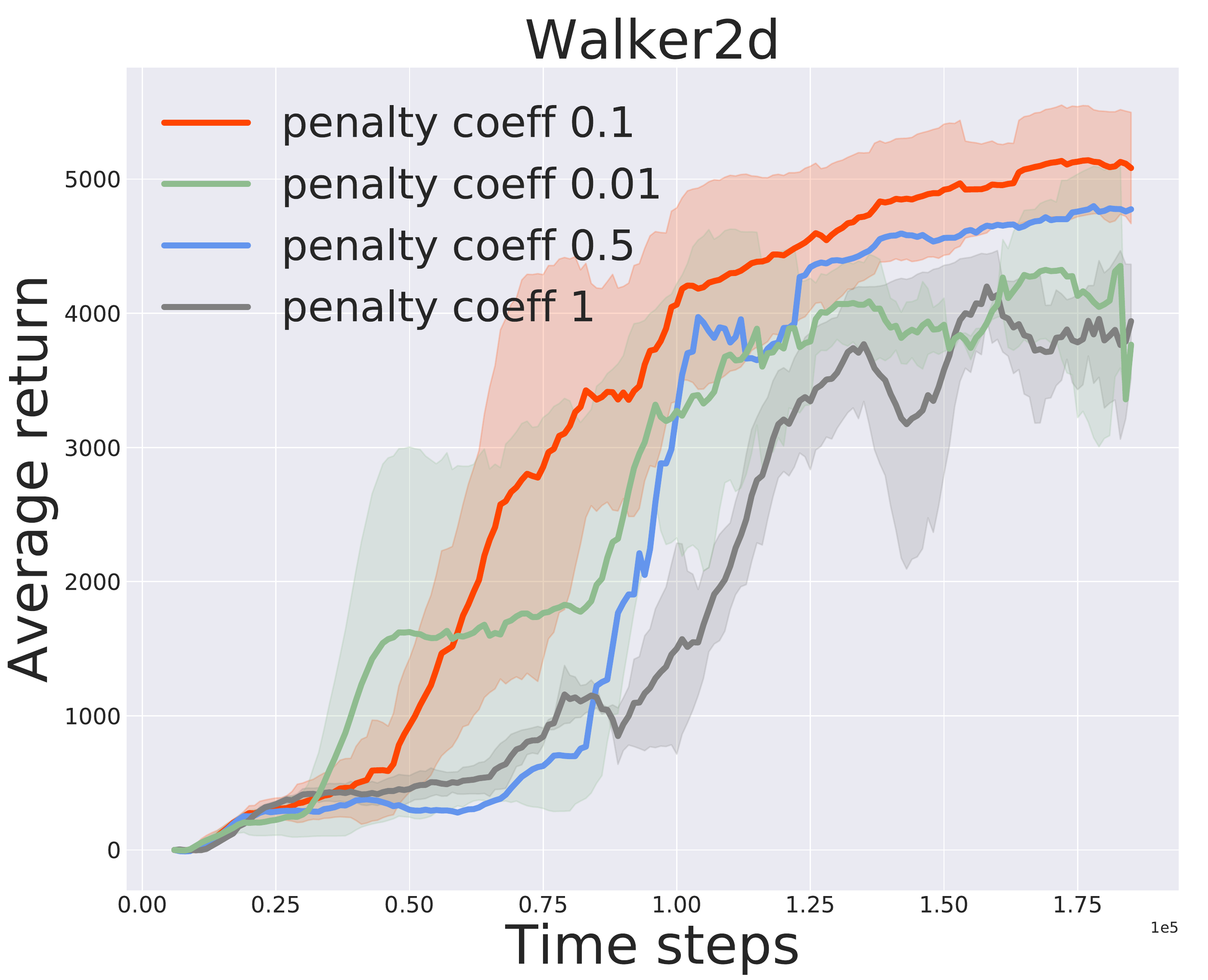}
    \caption{(Left two) Performance of CMBAC and CMBACUP on the Hopper and Humanoid environments. The results show that CMBACUP with the same penalty coefficient as the CMBACUP in Section 5.2 is sensitive to the environments. (Right)  Performance of CMBACUP with different penalty coefficients on Walker2d. The results demonstrate that CMBACUP is sensitive to the penalty coefficients in the same environment.}
    \label{fig:penalty_sensitivty}
\end{figure*}

\begin{table}[t]
	\caption{Common parameters used in the comparative evaluation and abalation study.}\label{hyper:shared}
	\centering 
	\vspace{0.3cm}
	\begin{tabular}{ l  l }
		\toprule
		Parameter  & Value  \\ 
		\midrule 
		environment steps per epoch & 1000\\
		policy updates per environment step & 20\\
		optimizer &Adam \\
		discount ($ \gamma $)&0.99\\
		probabilistic nn ensemble size & 7 (elite number 5)\\
        value network architecture & [256,256,256]\\
        policy network architecture & [256,256] \\
		\bottomrule
	\end{tabular}
\end{table}

\subsection{Hyperparameters}
    We use the same hyperparameter as that of MBPO \cite{janner2019mbpo} if possible. First, we list the common CMBAC parameters used in the comparative evaluation and ablation study in Table \ref{hyper:shared}. Second, we list additional CMBAC parameters, which are different across different environments. 
    
    \subsubsection{Probabilistic NN architecture} Four hidden layers of size 400 for Humanoid and four hidden layers of size 200 for the other five environments, which follows MBPO.
    
    \subsubsection{Model horizon} As \citet{janner2019mbpo}, $x\to y$ over epochs $a\to b$ denotes a thresholded linear function, i.e., at epoch $e$, $f(e) = \min(\max(x+\frac{e-a}{b-a}\cdot (y-x) , x),y)$. HalfCheetah and InvertedPendulum: 1; Hopper: $1\to15$  over epochs $20 \to 100$; The rest: $1\to 25$ over epochs $20\to 300$.
    
    \subsubsection{The number of probabilistic NN in a model $M$} As mentioned in Section 5, we select $M$ via grid search in $[1,2,3,4]$. The best number for Walker2d, Hopper, Humanoid, and the rest is 4,3,1, and 2, respectively. 
    
    \subsubsection{The number of dropped estimates $L$}
    As mentioned in Section 5, we select the number of dropped estimates via grid search in $[0,1,2]$. The best number for Walker2d and the rest is 0 and 1, respectively. 
    
\section{Additional Experimental Results}
    \subsection{Sensitivity analysis of CMBACUP}
    To test the sensitivity to the penalty coefficient of CMBACUP, we design two different kinds of experiments. 
    \begin{enumerate}
        \item First, we test the performance of CMBACUP with the same penalty coefficient in different environments. We select the penalty coefficient via grid search in $[0.01,0.1,0.5,1]$ on the Walker2d and Ant environments. We find that $0.1$ is the best. Note that Section 5.2 reports the best performance of CMBACUP. However, Figure \ref{fig:penalty_sensitivty} (left two) shows that CMBAC significantly outperforms CMBACUP on the Humanoid and Hopper environments. This result suggests that we need to carefully tune the penalty coefficient of CMBACUP in different environments to achieve high performance. However, CMBAC uses the same number of dropped estimates $L$ in all environments, except Walker2d. 
        
        \item Second, we test the performance of CMBACUP with different penalty coefficients in the same environment, i.e., Walker2d. The results in Figure \ref{fig:penalty_sensitivty} (right) demonstrate that CMBACUP is sensitive to the penalty coefficient. 
    \end{enumerate}
    Overall, these results suggest that CMBACUP tends to be sensitive to the penalty coefficient. 
    
    \subsection{Computational analysis} 
\begin{figure}
    \centering
    \includegraphics[width=0.45\textwidth]{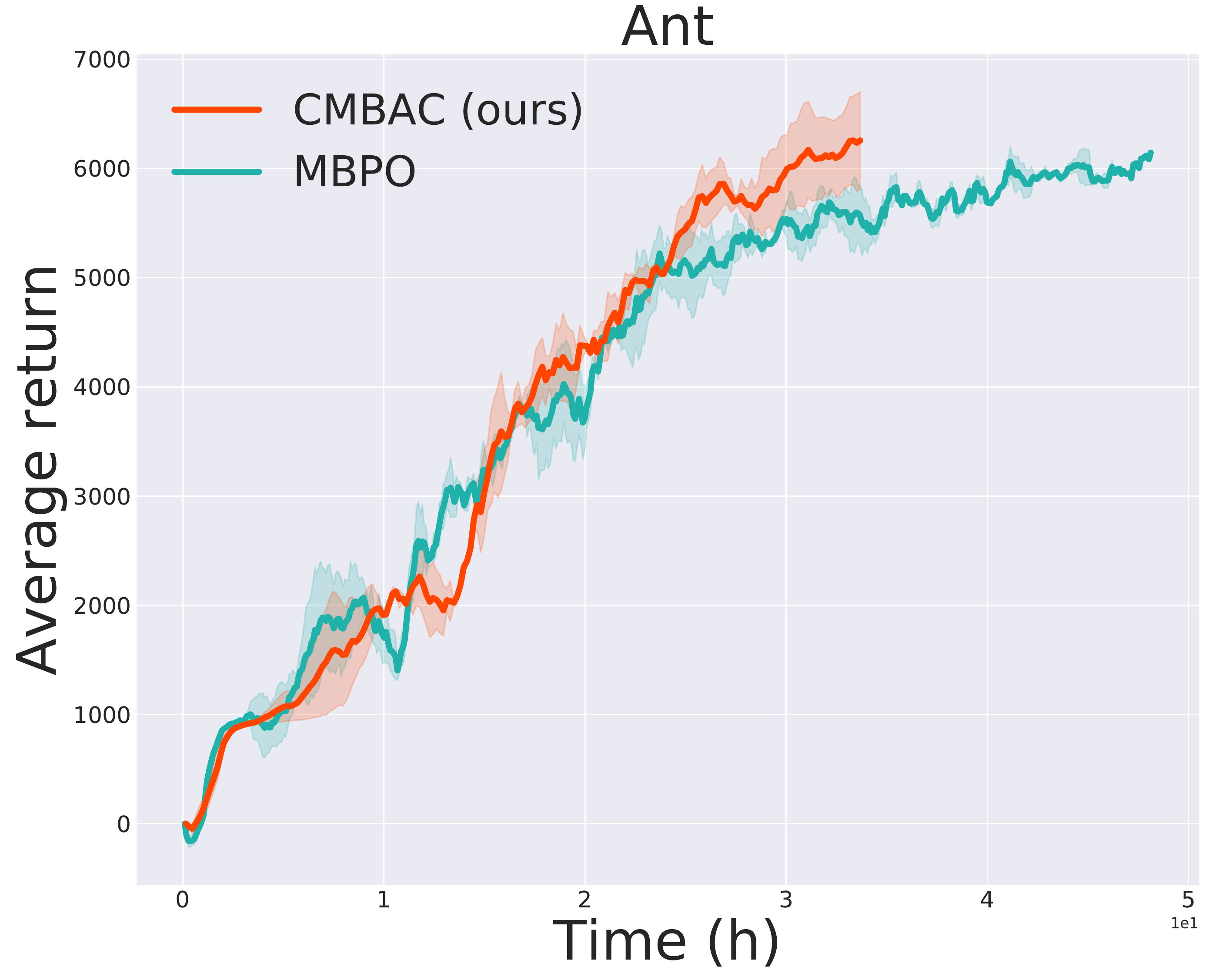}
    \caption{The wall clock time versus the average return on the Ant environment. The results show that CMBAC requires less computing time than MBPO to reach the maximum performance on Ant.}
    \label{fig:wall_clock}
\end{figure}

    We compare the computational complexity of CMBAC to MBPO. Specifically, we report the wall clock time versus the average return on the Ant environment when running the experiments on a single NVIDIA GeForce RTX 2080Ti GPU card in Figure \ref{fig:wall_clock}. The results in Figure \ref{fig:wall_clock} demonstrate that our method requires less computing time than MBPO to reach the maximum performance on the Ant environment.
    
    \subsection{Additional ablation study results}
    We provide additional ablation study results on the Hopper and HalfCheetah tasks in this subsection. 
    
    \subsubsection{Contribution of each component}
        The path from MBPO \cite{janner2019mbpo} to CMBAC comprises three modifications: Q-network size increase (Big), learning multiple estimates of the Q-value (LMEQ), and conservative policy optimization (CPO). B-MBPO is MBPO with an increased size of Q-networks (Big MBPO). The big Q-network has three layers with 512 neurons versus two layers of 256 neurons in MBPO. Note that the policy network size does not change. B-LMEQ-MBPO is B-MBPO with learning multiple estimates of the Q-value to capture the uncertainty of the Q-value function. B-LMEQ-MBPO uses the average of all the estimates to optimize the policy. CMBAC is B-LMEQ-MBPO with conservative policy optimization, i.e., it uses the average of the bottom-k estimates to optimize the policy. 
        
\begin{figure}[t]
    \centering
    \includegraphics[width=0.48\textwidth]{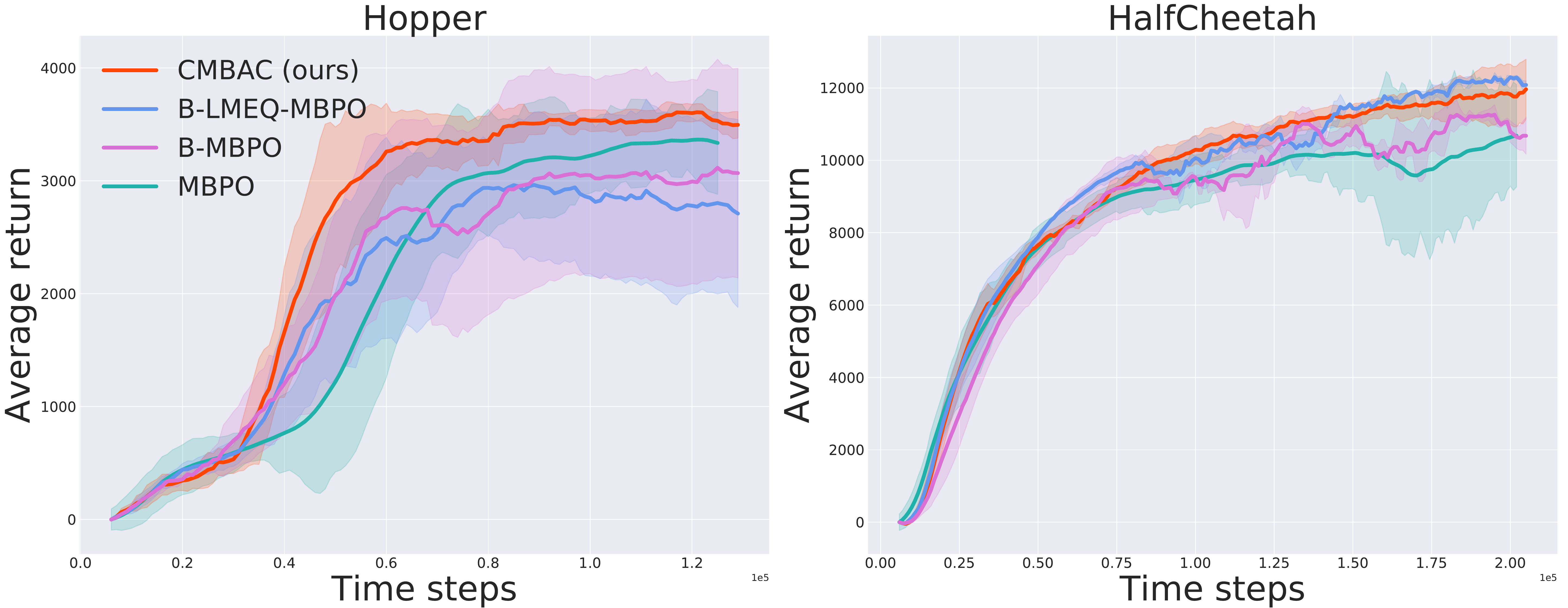}
    \caption{Performance of CMBAC and its ablations in the Hopper and HalfCheetah environments. The results demonstrate that each component of CMBAC is important for performance improvement on Hopper.} 
    \label{fig:more_component}
\end{figure}

        Figure \ref{fig:more_component} shows that both LMEQ and CPO are significant for performance improvement in the Hopper environment. On HalfCheetah, B-LMEQ-MBPO achieves comparable performance to CMBAC. We find that CMBAC achieves more stable performance than B-LMEQ-MBPO in the early stage of training, but the asymptotic performance of CMBAC is a little lower than B-LMEQ-MBPO. One possible reason is that the agent can be more optimistic when the model becomes more accurate. 
    
    \subsubsection{Sensitivity analysis}
\begin{figure}[t]
    \centering
    \includegraphics[width=0.48\textwidth]{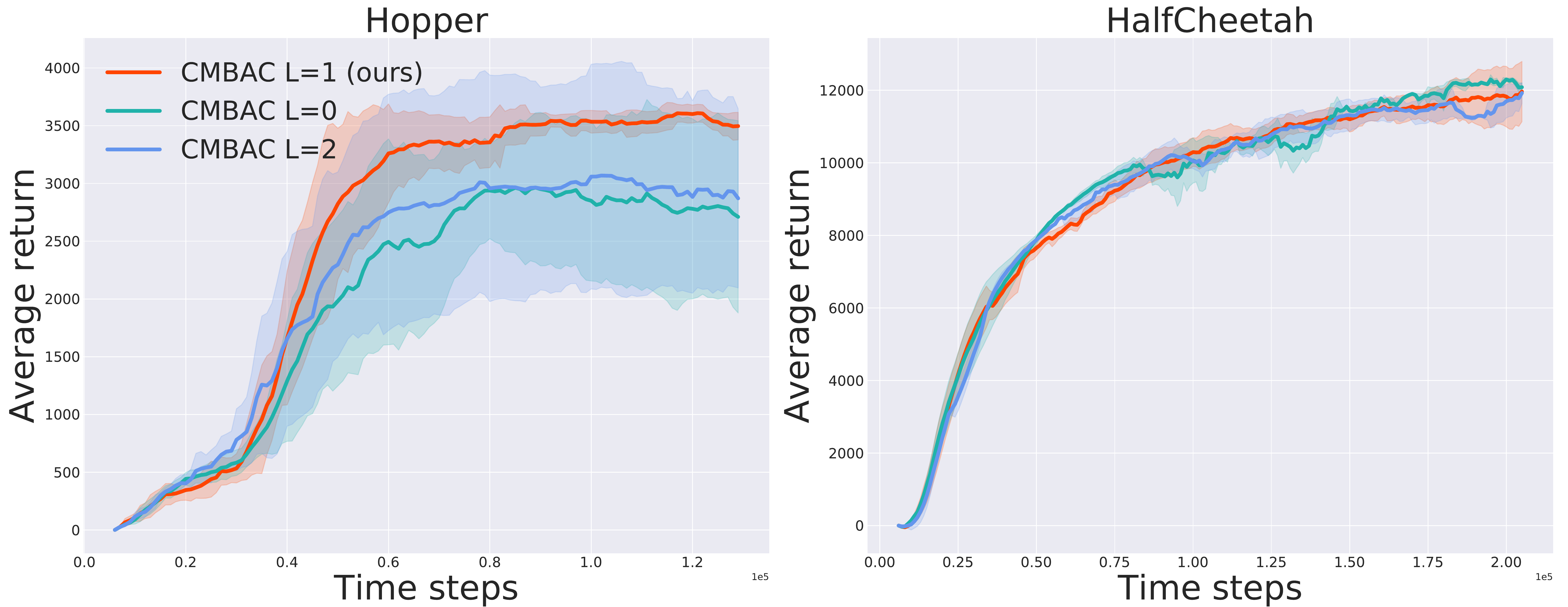}
    \caption{Varying the number of dropped estimates $L$. The results show that $L$ is essential on Hopper, and $L=1$ provides significant performance improvement on Hopper.} 
    \label{fig:more_drop}
\end{figure}

        On Hopper and HalfCheetah, we vary the number of dropped estimates for each ensemble multi-head Q-network $L\in\{0,1,2\}$. The total number of estimates dropped is $2L$.
        Figure \ref{fig:more_drop} suggests that dropping $L$ topmost estimates provides a performance improvement in the Hopper environment but does not further improve the performance in the HalfCheetah environment.

\section{Additional Details of CMBAC}

\subsection{Multi-head Q-network architecture}
\begin{figure}[t]
    \centering
    \includegraphics[width=0.45\textwidth]{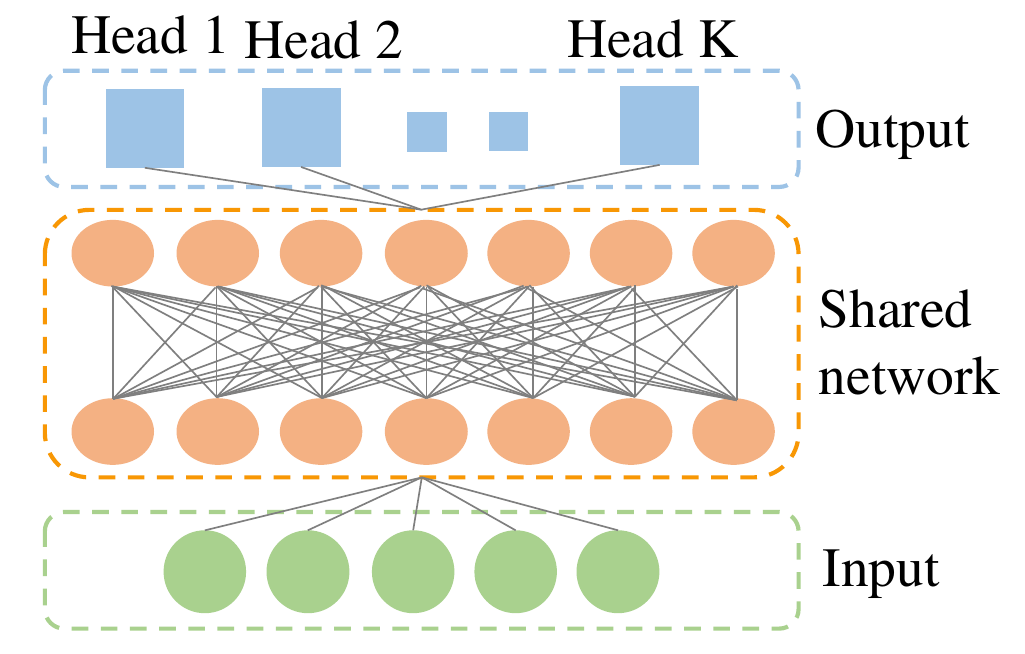}
    \caption{An illustration of the multi-head Q-network architecture. The multi-head Q-network outputs K ``heads''.}
    \label{fig:multi-head q}
\end{figure}

   To approximate the posterior distribution over Q-values, we use a multi-head Q-network similar to \citet{bootstrap_dqn}. As shown in Figure \ref{fig:multi-head q}, the multi-head q-network outputs $K$ different values. In our experiment, each ``head'' provides an estimate of the Q-value. \citet{bootstrap_dqn} show that this network architecture can provide significant computational advantages at the cost of lower diversity between heads.  

\subsection{Details of CMBAC variants}
    We present a detailed description of the proposed CMBAC variants in Section 5.2.
    
    \subsubsection{MBPOEQ} As CMBAC, MBPOEQ uses two multi-head Q-networks, denoted by $\{Q^1_{i}, Q^2_{i}\}_{i=1}^{K}$. Given $(s,a,s^{\prime})$ sampled from $\mathcal{D}_{\text{model}}$, the target of each ``head'' is given by
    \begin{align}
        y(s,a) = r(s,a) + \gamma \min_{j=1,2} (\frac{1}{K} \sum_{i=1}^{K} Q^j_i(s^{\prime},a^{\prime})),
    \end{align}
    where $a^{\prime}\sim \pi(\cdot|s^{\prime})$.  
    
    \subsubsection{CMBACUP} We define the standard deviation of the estimates given by the multi-head Q-network as our estimated uncertainty $u_{Q}(s,a)$. Given state $s$ and action $a$, we use the uncertainty $u_{Q}(s,a)$ as a penalty instead of dropping several topmost estimates. 
    
    \subsubsection{MINCMBAC} Given the state $s$ and action $a$, MINCMBAC uses the minimum estimate of all possible estimates to optimize the policy. As discussed in Section \ref{alg:dis}, MINCMBAC is different from robust policy optimization. 
    
    \subsubsection{REDQ-CMBAC} Both CMBAC and REDQ-CMBAC learn two multi-head Q-networks and obtain a conservative estimate by dropping several topmost estimates for each multi-head Q-network. CMBAC then uses the minimum of the two conservative estimates to optimize the policy similar to the clipped double Q-learning trick \cite{td3}. In contrast, REDQ-CMBAC uses the average of the two conservative estimates instead of the minimum to optimize the policy as \citet{redq}.  

\begin{figure*}[t]
    \centering
    \includegraphics[width=0.90\textwidth]{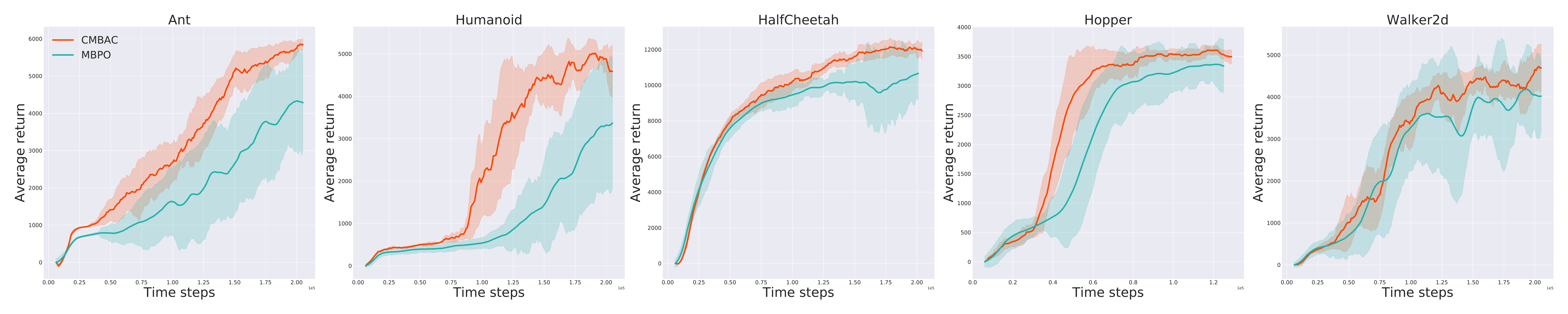}
    \caption{Performance of CMBAC and MBPO on five benchmark tasks.}
    \label{fig:eva_new_para}
\end{figure*}

\begin{figure}[t]
    \centering
    \includegraphics[width=0.48\textwidth]{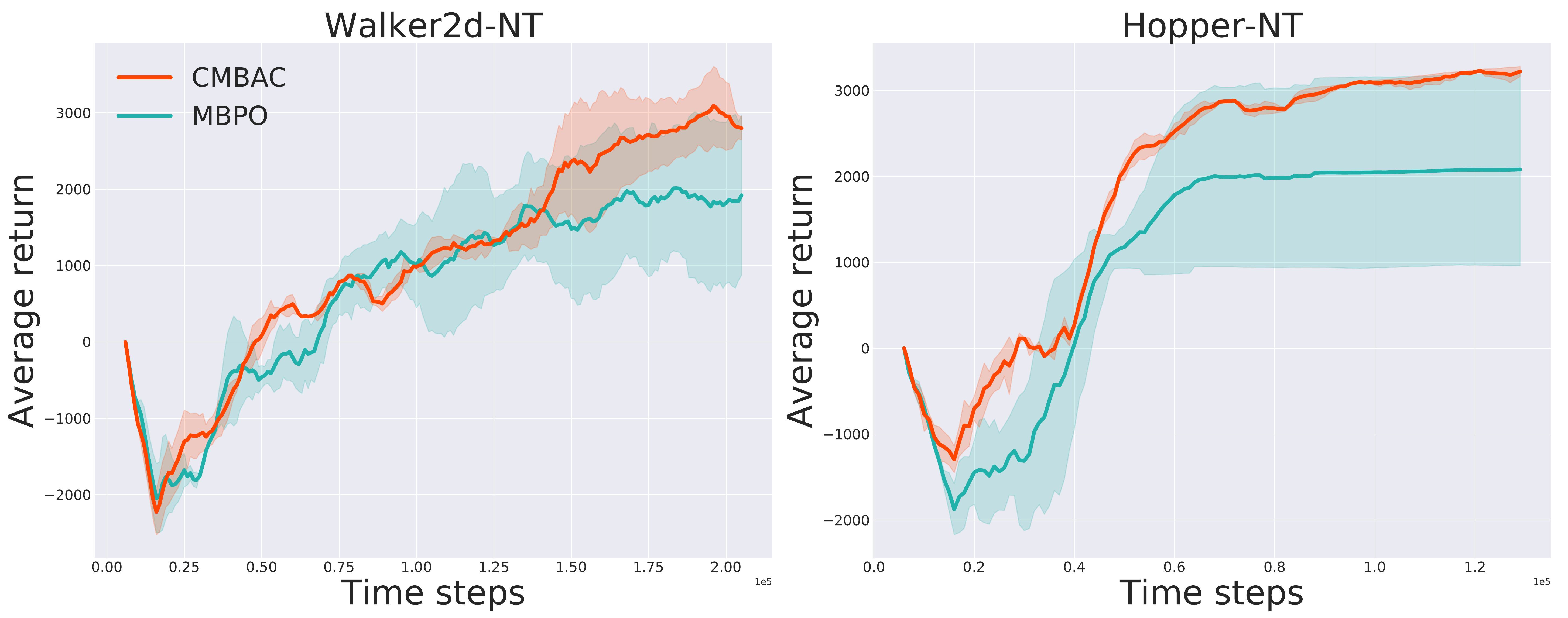}
    \caption{Results in two new
challenging environments.}
    \label{fig:eva_new_env}
\end{figure}

\subsection{Details of visualization experiments}
We present detailed experimental settings and analyses of the visualization experiments in Section 5.3. 

\subsubsection{Uncertainty estimation}
    In Section 5.3, we find that the cumulative discounted sum of prediction errors can hardly approximate the errors of the Q-function. Here we provide possible reasons. One possible reason is due to compounding model errors. Similar to \citet{mopo}, we approximate the prediction error at each step by the uncertainty estimated using Eq \ref{eq:mopo_uncertainty}.  We compute the cumulative discounted sum of prediction errors by generating a 1000-step rollout in the model. In this case, the cumulative discounted sum of prediction errors can be large due to compounding model errors. Another possible reason is that the uncertainty quantification in \citet{mopo} is inappropriate for the online setting, as it is designed for the offline setting.
    
\subsubsection{Conservative policy optimization}
    We provide details of the 2D point environment and the simplified CMBAC. 
    
    \textbf{2D point environment} Our 2D point environments is similar to that defined by \citet{mb_mpo}. The state space is $[-2,2]^2$. The action space is $[-1,1]^2$. The initial state distribution is an uniform distribution over the state sapce. The transition function is defined by
    \begin{align*}
        f(s,a) = \text{clip}(s+a, -2, 2),
    \end{align*}
    where clip represents a pointwise clipping of $s+a$ to bound the next state. The goal the agent is to reach the goal state, i.e., $g=(0,0)$. 
    Following \citet{sql}, we define the reward function by
    \begin{align*}
        r(s,a) = c \cdot \exp(-\frac{\|f(s,a)-g\|^2}{\alpha}),
    \end{align*}
    where $c=0.05$ and $\alpha=5$. The horizon is $50$. 
    
    \textbf{A simplified CMBAC} To disentangle different components in CMBAC, we implement a simplified CMBAC, named SCMBAC, in this visualization experiment. To analyze the errors of the true Q-value in the model-based setting, SCMBAC removes the entropy and does not use the clipped double Q-learning trick.

\section{Additional results of CMBAC}

\subsection{Evaluation with unified hyperparameters}
    For high-dimensional tasks, i.e., Humanoid and Ant, we use $M = 1$. For the rest, we use $M = 3$. For all tasks, we
    use $L = 1$. Figure \ref{fig:eva_new_para} shows that CMBAC using the unified
    hyperparameters significantly outperforms MBPO on several challenging tasks.

\subsection{Results on two additional environments}
    We conduct experiments using the unified hyperparameters in two additional challenging benchmark environments, i.e., Walker2d-NT and Hopper-NT, which are different from Walker2d and Hopper in the terminal functions and reward functions \cite{benchmark}. Figure \ref{fig:eva_new_env} shows that CMBAC consistently outperforms MBPO on Walker2d-NT and Hopper-NT. 

\end{document}